\documentclass[conference]{IEEEtran}

\usepackage{amsmath}
\usepackage[pdftex]{graphicx}
\usepackage{algorithm}
\usepackage{algorithmic}
\usepackage{subfigure}
\usepackage{booktabs}
\usepackage{array}
\usepackage{caption}

\newcommand{\nop}[1]{}

\newcommand{\seqfig}[3]{
	{\begin{figure}[!t]
	\centering
	\includegraphics[width=#2\linewidth]{seqfigures/#1}
	\caption{#3}
	\label{seqfig:#1}
	\end{figure}}
}

\begin{document}

\title{Recomposition vs. Prediction: A Novel Anomaly Detection for Discrete Events Based On Autoencoder}

\author{
\IEEEauthorblockN{Lun-Pin Yuan}
\IEEEauthorblockA{Penn State University\\
lunpin@psu.edu}
\and
\IEEEauthorblockN{Peng Liu}
\IEEEauthorblockA{Penn State University\\
pliu@ist.psu.edu}
\and 
\IEEEauthorblockN{Sencun Zhu}
\IEEEauthorblockA{Penn State University\\
sxz16@psu.edu}
}

% make the title area
\maketitle

\begin{abstract}
One of the most challenging problems in the field of intrusion detection is \textit{anomaly detection for discrete event logs}.  While most earlier work focused on applying unsupervised learning upon engineered features, most recent work has started to resolve this challenge by applying deep learning methodology to abstraction of discrete event entries.  Inspired by natural language processing, LSTM-based anomaly detection models were proposed.  They try to predict upcoming events, and raise an anomaly alert when a prediction fails to meet a certain criterion.  However, such a \textit{predict-next-event} methodology has a fundamental limitation: \textit{event predictions may not be able to fully exploit the distinctive characteristics of sequences}.  This limitation leads to high false positives (FPs) and high false negatives (FNs).  It is also critical to examine the structure of sequences and the bi-directional causality among individual events.  To this end, we propose a new methodology: \textit{Recomposing event sequences as anomaly detection}.  We propose DabLog, a LSTM-based \textit{\textbf{D}eep \textbf{A}utoencoder-\textbf{B}ased anomaly detection method for discrete event \textbf{Log}s}.  The fundamental difference is that, rather than predicting upcoming events, our approach determines whether a sequence is normal or abnormal by analyzing (encoding) and reconstructing (decoding) the given sequence. Our evaluation results show that our new methodology can significantly reduce the numbers of FPs and FNs, hence achieving a higher $F_1$ score.  
\end{abstract}

    {\section{Introduction}

One of the most challenging problems in the field of intrusion detection is \textit{anomaly detection for discrete event logs}.  Researchers have been trying to resolve this challenge for two decades, and most work have focused on applying unsupervised learning upon engineered features from normal data, assuming unforeseen anomalies do not follow the learned normal patterns (e.g.,~\cite{svddc, nlsalog16, pca01, pca02, log2vec, nlsalog20, lifelong15, lifelong26, liuliu1, liuliu2}).  Recently, solving this challenge with deep learning has gained a substantial amount of traction in the security community (e.g.,~\cite{deeplog, nlsalog9, nlsalog12, habad, nlsalog}), partially due to the unique advantages of deep learning in natural language processing.  Researchers have applied related language-processing methodologies to anomaly detection for discrete event logs by treating discrete events as words and logs as sentences, as if linguistic causality exists in the security logs.  The main benefit of this approach over machine learning upon engineered features is that detail domain knowledge, complex feature extraction, and costly human interference are no longer required  (e.g.,~\cite{liuliu2, kitsune22, kitsune23, liuliu3, spell}).  

Inspired by natural language processing, Long Short Term Memory (LSTM)~\cite{lstm} based anomaly detection models (e.g.,~\cite{deeplog, nlsalog9, nlsalog12, habad, nlsalog}) were proposed.  These models try to predict upcoming log events, and they raise an anomaly alert when a prediction fails to meet a certain criterion.  However, we found that the widely-adopted methodology \textit{``using an LSTM-based model in predicting next events''} has a fundamental limitation: \textit{\textbf{event predictions may not be able to fully exploit the distinctive characteristics of sequences}}.  To be specific, event-prediction methodology assumes the distribution of an event is affected only by the prior events before it (e.g., when a model sees an \textit{open} file-operation, it can guess such \textit{open} operation is followed by \textit{read} operations); however, the distribution can also be affected by later events (e.g., when a model sees a \textit{read} operation, it should examine whether it has seen any \textit{open} operation) or no events whatsoever (e.g., an event may have nothing to do with the other events).  Therefore, an anomaly detection method should also look deeper into the sequential structure and the bi-directional causality among events.  Because of this limitation, the widely adopted methodology could lead to numerous false positives (FPs) and false negatives (FNs).  

Examples why the methodology could lead to FPs and FNs are illustrated as follows.  For FP example, consider a normal sequence of file operations $[\textit{open~A}, \textit{read~A}, \textit{read~A}]$ and an upcoming event \textit{open~B}.  By seeing just the first few operations, a predictor-based anomaly detection model may guess the upcoming event to be \textit{read~A}, because (1) it is one of the most frequent events that follow \textit{open~A} in the training dataset while \textit{open~B} is less frequent, and (2) the first few operations does not enclose prior knowledge which indicates \textit{B} will be soon opened; consequently, the predictor-based model may wrongly report the sequence as abnormal, although in reality it is also normal but less common. 
%This is similar to the \textit{cold-start} issue in a recommendation system where a new user is typically recommended with popular items among all existing users;
The fundamental issue is that, when little necessary knowledge is available in a sequence (regardless of sequence length), predictor-based anomaly detection always has to make bold guesses. For FN example, consider an abnormal sequence of file operations $[\textit{read~A}, \textit{read~A}, \textit{close~A}]$ and an upcoming event \textit{read~A}.  If the predictor-based model does not examine whether there is any \textit{open~A} before the upcoming \textit{read~A}, it may consider this sequence normal,
%if the model correctly predicts \textit{read A}, 
hence a false negative.

To address the fundamental limitation of \textit{not being able to fully exploit the distinctive characteristics of sequences}, we propose a different methodology: \textit{\textbf{using an LSTM autoencoder in recomposing sequences}}.  Compared to the existing methodology, the fundamental difference is that our LSTM autoencoder determines whether a sequence is normal or abnormal by analyzing (encoding) and reconstructing (decoding) the given sequence rather than predicting upcoming individual events. The intuition is that an anomaly detection method should see a sequence as an atomic instance, and it should examine the structure of the sequence as well as the bi-directional causality among the events.  Hence, our anomaly detection model can detect not only sequences that include unseen or rare events, but also structurally abnormal sequences. 
Note that, our model is more than a standard autoencoder which reconstructs input input vectors.  To work with discrete events, our solution is designed as an \textit{embed-encode-decode-classify-critic} model.

In this work, we propose DabLog, a \textit{\textbf{D}eep \textbf{A}utoencoder-\textbf{B}ased anomaly detection method for discrete event \textbf{Log}s}.  DabLog aims to provide an anomaly detection function $\mathcal{AD}: \mathcal{S} \rightarrow \{\text{normal},$ $\text{abnormal}\}$.  DabLog consists of four major components (Figure~\ref{seqfig:model.pdf}): an embedding layer, a deep LSTM autoencoder, an event classifier, and an anomaly critic.  Our evaluation results show that the new methodology can significantly reduce the number of FPs, while achieving a better $F_1$ score.  
Compared to our predictor-based baseline model, DabLog reports 1,790 less FPs but 1,982 more TPs in our evaluation upon HDFS console logs with 101 distinct events, and DabLog reports 2,419 less FPs with trade-off 83 less TPs in our evaluation upon traffic logs with 706 distinct events.  Specifically, we make the following contributions.

\begin{enumerate}
\item Through in-depth FP and FN case studies, we discover a fundamental limitation of predictor-based models.  We resolve this limitation by proposing a deep autoencoder-based anomaly detection method for discrete event logs.
\item We evaluate DabLog upon HDFS console-log dataset, and our results show that DabLog outperforms our re-implemented predictor-based baseline model in terms of $F_1$ score.  DabLog achieves $97.18\%$ $F_1$ scores in evaluation with 101 distinct events, while our baseline model achieves only $87.32\%$ $F_1$ scores.
\item To the best of our knowledge, we are the first to show that autoencoders can effectively serve the purpose of detecting \textit{time-sensitive} anomalies in discrete event logs with contexts that involves more distinct events, while the common practice is to apply predictors to time-sensitive data with fewer distinct events and to apply autoencoders to time-insensitive data.  
\end{enumerate}

The rest of this paper is organized as follows. 
Section~\ref{seqsec:relatedwork} outlines related work in the anomaly detection literature.  Section~\ref{seqsec:background} introduces background knowledge to encoder-decoder networks.  Section~\ref{seqsec:motivation} states our motivation and a motivating example.  Section~\ref{seqsec:dablog} details our DabLog design.
Section~\ref{seqsec:evaluation} shows our evaluation results and case studies.  Section~\ref{seqsec:discussion} discusses a few limitations and future work.  Lastly, Section~\ref{seqsec:conclusion} concludes this paper.

}

    {\section{Related Work} \label{seqsec:relatedwork}

Most anomaly detection methods are \textit{zero-positive} machine learning models that are trained by only normal (i.e., negative) data and then used in testing whether observation data is normal or abnormal, assuming unforeseen anomalies do not follow the learned normal patterns.  For example, Kenaza et al.~\cite{svddc} integrated supports vector data description and clustering algorithms, and Liuq  et al.~\cite{nlsalog16} integrated K-prototype clustering and k-NN classification algorithms to detect anomalous data points, assuming anomalies are rare or accidental events.  When prior domain knowledge is available for linking causal or dependency relations among subjects and objects and operations, graph-based anomaly detection methods (such as Elicit~\cite{pca02}, Log2Vec~\cite{log2vec}, Oprea et al.~\cite{nlsalog20}) could be powerful. 
%For example, Maloof and Stephens proposed Elicit~\cite{pca02}, which leverages Bayesian inference network, Liu et al. proposed Log2Vec~\cite{log2vec}, which leverages clustering algorithms upon graphs, and Oprea et al.~\cite{nlsalog20} proposed a model which leverages a belief propagation algorithm.  
When little prior domain knowledge is available, Principal Component Analysis (PCA) based anomaly detection methods (for example, Hu et al.~\cite{pca01} proposed an anomaly detection model for heterogeneous logs using singular value decomposition) could be powerful.  Oppositions to zero-positive anomaly detection are semi-supervised or online learning anomaly detection, in which some anomalies will be available over time~\cite{lifelong}.

Autoencoder framework is another PCA approach that is widely used in anomaly detection.  Briefly speaking, a typical autoencoder-based anomaly detection method learns how to reconstruct normal data, and it detects anomalies by checking whether the reconstruction error of a data point has exceeded a threshold.  To detect anomalies, Zong et al.~\cite{lifelong48} proposed deep autoencoding Gaussian mixture models, Chiba et al.~\cite{chiba} proposed autoencoders with back propagation, Sakurada and Yairi~\cite{lifelong31} proposed autoencoders with nonlinear dimensionality reduction, Lu et al. proposed MC-AEN~\cite{exploitembed} which is an autoencoder which is constrained by embedding manifold learning, Nguyen et al. proposed GEE~\cite{gee} which is a variational autoencoder with gradient-based anomaly explanation, Wang et al. proposed adVAE~\cite{advae} which is a self-adversarial variational autoencoder with Gaussian anomaly prior assumption, Alam et al. proposed AutoPerf~\cite{lifelong15} which is an ensemble of autoencoders accompanied by K-mean clustering algorithm, Mirsky et al. proposed Kitsune~\cite{lifelong26} which is an ensemble of lightweight autoencoders, Liu et al.~\cite{liuliu1, liuliu2} proposed an ensemble of autoencoders for multi-sourced heterogeneous logs, and Chalapathy et al.~\cite{rcae} and Zhou et al.~\cite{lifelong47} proposed robust autoencoders.

The above anomaly detection methods only work with \textbf{\textit{time-insensitive}} data (i.e., each data point is independent of the other data points).  To work with \textbf{\textit{time-sensitive}} data (i.e., dependencies exist among data points), researchers have leveraged Long Short-Term Memory (LSTM)~\cite{lstm} in building anomaly detection models.  LSTM has been widely used in learning sequences, and LSTM-based deep learning has been widely used to extract patterns from massive data.  Since most cyber operations are sequential (e.g., as in timestamped audit logs), LSTM-based deep learning has great potential in serving anomaly detection applications.  Inspired by natural language processing, Deeplog~\cite{deeplog}, Brown et al.~\cite{nlsalog9}, DReAM~\cite{nlsalog12}, HAbAD~\cite{habad}, and nLSALog~\cite{nlsalog} were proposed to build LSTM-based multi-class classifier in order to predict future log entries.  We summarize these LSTM-based methods in the next section; for other anomaly detection methods, surveys and comparisons can be found in~\cite{deepsurvey1, deepsurvey2, deepsurvey3, deepsurvey4, survey1, nlsalog10, nlsalog25}.
}

    {

\section{Background Knowledge} \label{seqsec:background}

\seqfig{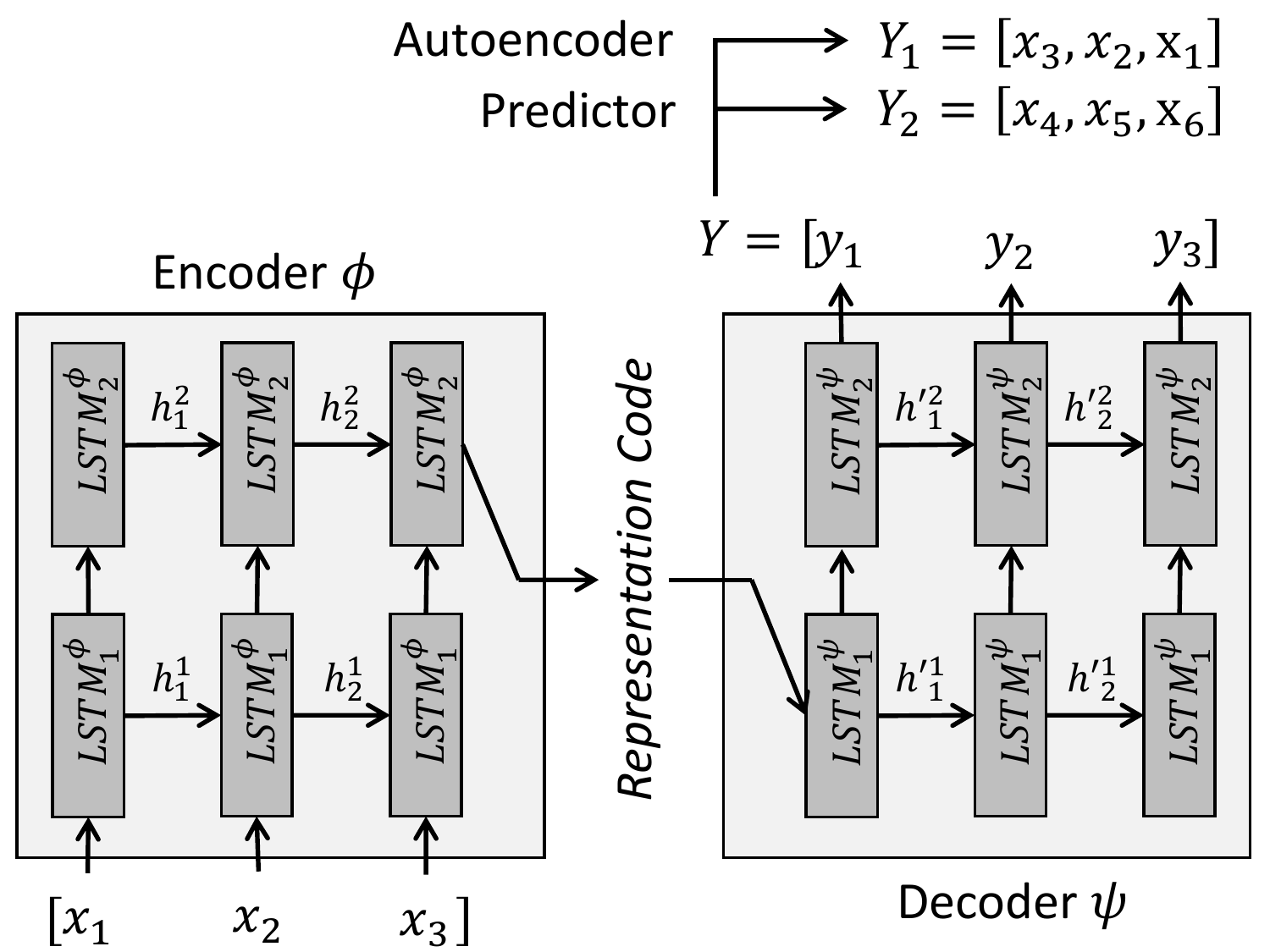}{0.9}{Deep LSTM Encoder-Decoder Network}

To understand our approach, some background knowledge on \textbf{\textit{Deep LSTM Encoder-Decoder Network}} is essential.  Cho et al.~\cite{lstmae1} proposed an LSTM encoder-decoder network for statistical machine translation.
%(note that an LSTM encoder-decoder network is not necessarily an LSTM autoencoder network).
Both encoder and decoder are recurrent networks.  An encoder $\phi$ takes a variable-length input sequence $X=[x_1, x_2, x_3, \dots, x_T ]$ of length $T$ and generates a brief fixed-length \textbf{\textit{representation code}} (also commonly referred to as the \textit{representation} or the \textit{code}) of $X$, and the encoding operation is denoted as $\text{code}=\phi(X)$).  A decoder $\psi$ then takes the representation code and generates a variable-length target sequence $Y=[y_1, y_2, y_3, \dots, y_{\mathcal{T}} ]$ of length $\mathcal{T}$, and the decoding operation is denoted as $Y=\psi(\text{code})=(\psi \circ \phi)(X)$.  Depending on the application, $X$ and $Y$ may have different lengths.  Srivastava et al.~\cite{lstmae2} summarized three types of LSTM encoder-decoder networks for unsupervised learning models.% of video representations.

\begin{enumerate}
\item \textbf{\textit{Autoencoder}}:  The goal of an autoencoder is to enclose into the representation code all needed to \textit{reconstruct the same sequence}.  An autoencoder takes an input sequence $X=[x_1, x_2, x_3, \dots, x_T]$ and tries to reconstruct the target sequence $\hat{Y_1}=[x_T, x_{T-1}, x_{T-2}, \dots, x_1]$.  Note that the target sequence is in the reverse order, as if the encoder recurrently encodes (pushes) $x_t$ into the representation code, whereas the decoder recurrently decodes (pops) $x_t$ from the code.
\item \textbf{\textit{Predictor}}:  The goal of a predictor is to \textit{predict future sequence} based on what it has observed.  The representation code plays the role of an internal hidden state. A predictor takes the input sequence $X=[x_1, x_2, x_3, \dots, x_T]$ and tries to predict the target sequence $\hat{Y_2}=[x_{T+1}, x_{T+2}, x_{T+3}, \dots, x_{T+\mathcal{T}}]$.  If $\mathcal{T}=1$, then it is a single-event predictor.
\item \textbf{\textit{Composite}}:  Merging the above two models, a composite model tries to reconstruct-predict $\hat{Y_3}=[\hat{Y_1}, \hat{Y_2}]$.
\end{enumerate}

Each LSTM encoder-decoder network listed above can be \textbf{\textit{conditional}} or \textbf{\textit{unconditional}}, depending on whether the true $\hat{y}_{\tau}$ (\textit{condition}) is provided to the decoder (as an additional input) when the decoder tries to decode $y_{\tau+1}$.  In an autoencoder, $\hat{y}_\tau=x_{T-\tau+1}$, whereas in a predictor $\hat{y}_\tau=x_{T+\tau}$.

The time-sensitive anomaly detection models for discrete event logs, mentioned in Section~\ref{seqsec:relatedwork}, are predictors.  These predictors typically consider an upcoming event $x_{T+\tau}$ normal if the probability  $\text{Pr}(x_{T+\tau} | x_1, x_2, \dots, x_{T+\tau-1})$ is within a threshold (or alternatively $x_{T+\tau}$ is within the top-$N$ prediction); otherwise abnormal. Specifically, DeepLog~\cite{deeplog} leverages a two-layer LSTM network that works on one-hot representation of log entries. 
%Brown et al.~\cite{nlsalog9} proposed 
Brown et al.~\cite{nlsalog9} leverages bidirectional LSTM, word embedding, and five attention mechanisms.  %Ezeme et al. proposed 
Both HAbAD~\cite{habad} and 
%Ezeme et al. proposed 
DReAM~\cite{nlsalog12} build an embed-encoder-attention-decoder framework. 
%Yang et al. proposed 
nLSALog~\cite{nlsalog} leverages n-layer stacked LSTM, embedding layer, and self-attention mechanism. Some of the above models further incorporate an \textit{embedding layer} (here embedding is a learned representation of log entries) in their encoders in order to include correlation among log entries, and some incorporate an \textit{attention layer} (here attention is an aggregated state of hidden states from each time-step or each neuron) in their decoders in order to improve prediction accuracy.

Predictors and autocoders have been extensively studied for time-sensitive anomaly detection, %and autoencoders have been extensively studied 
and for time-insensitive anomaly detection, respectively. However, to our best knowledge, whether autoencoders can serve time-sensitive anomaly detection have not yet been investigated until this work.  Conceptually, an LSTM autoencoder is essentially trying to learn the identity function of the input data distribution. Such identity function will definitely fail to fit every input data %(e.g., fail to simply copy its inputs to outputs) 
because, at high-level, there are only a fixed number of hidden units at each layer (in both encoder and decoder) and thus very unlikely they can learn everything needed for reconstruction. Moreover, hidden states (e.g., $h^i_j$ and $h'^i_j$ in Figure~\ref{seqfig:autoencoder.pdf}) and representation codes are too small to enclose detail information of the input data. Based on these constraints, autoencoders are forced to learn more meaningful concepts and relationships inside the input data.  Trained with only normal input, autoencoders can be used in detecting anomalies in case of poor reconstruction.

}

    {\section{Motivation} \label{seqsec:motivation}

We define an anomaly detection function for discrete events as $\mathcal{AD}: \mathcal{S} \rightarrow \{\text{normal}, \text{abnormal}\}$, where a sequence of events $S=[e_t | 1 \leq t \leq T] \in \mathcal{S}$ is essentially a set of relevant events (e.g., events of the same subject) sorted by timestamps (e.g., from past to present). Each discrete event $e_t$ is represented by a distinct event key $k_i \in \mathcal{K}$, which is a string template.  Distinct event keys are referred to as \textit{logkey} in DeepLog~\cite{deeplog}, \textit{log template} in nLSALog~\cite{nlsalog}, and \textit{discrete keys} in Du et al.~\cite{lifelong}'s work. 

Among the aforementioned related work, we find DeepLog~\cite{deeplog} and nLSALog~\cite{nlsalog} representative of predictor-based anomaly detection methods.  They are similar predictors that predict only single upcoming event $e_{T+1}$ for each sliding-window subsequence $s_T=[e_t | \text{max} (1, T - 9) \leq t \leq T]$, whose window size $|s_T|$ is at most ten (we refer this configuration to as $\text{seqlen}=10$). They consider $S$ anomalous if any prediction $e_{T+1} \in S$ is not an instance of event key $k_j \in \mathcal{K}$ in its top-$9$ predictions out of $28$ event keys, or equivalently top-$32\%$ predictions. Both methods were evaluated upon the same HDFS dataset~\cite{sosp2009, icdm2009} and seemed promising based on  $\text{accuracy}=(TP+TN)/(TP+FN+FP+TN)$. . 

Single-event prediction, however, is not an ideal solution for sequence-based anomaly detection $\mathcal{AD}: \mathcal{S} \rightarrow \{\text{normal}, \text{abnormal}\}$.  Typical anomaly detection methods are based on the variance of an instance, or equivalently the error from particular expectation of an instance.  In our context, the instances are the sequences $S \in \mathcal{S}$, and hence intuitively we should consider an individual sequence $S$ as an atomic instance.  Yet, by examining whether an individual event $e_{T+1} \in S$ is in top-$N$ expectation based on prior events, %$s_T$,
single-event prediction obviously considers the individual event $e_{T+1}$ as an atomic instance.  As such, it seems to us that single-event prediction is more like anomalous event detection, or somewhat rare event detection considering their configuration setup.  
The problem is twofold.  On one hand, a rare event does not necessarily make the event itself or the sequence abnormal, and hence wrongly reporting rare events as abnormal will cause more FPs (a case study is provided below). 
%That said, without any doubt, the existence of an unseen or abnormal event makes the corresponding sequence abnormal, but 
On the other hand, the absence of abnormal events does not necessarily makes the sequence normal. That is, a sequence without abnormal events can still be structurally abnormal; therefore, not checking sequential structure may cause more FNs (a case study is provided in Section~\ref{seqsec:evaluation}).  Based on the above observations, it is important to examine the structure of a sequence (and even to reconstruct it) as well as its bi-directional causality, and in fact by doing so one can expect significantly less FPs and FNs (details are in Section~\ref{seqsec:evaluation}).

\subsection{Motivating Example: FP Case Study}

We are particularly interested in how predictor-based approach can be applied to scenarios where finer grained prediction is required and more event keys are involved.  As opposed to the criterion of top-$32\%$ predictions out of $|\mathcal{K}|=28$ keys, our criterion top-$9\%$ out of $|\mathcal{K}_1|=101$ keys is more reasonable, and the reasons are detailed in the next subsection. We re-implement a predictor-based anomaly detection model (referred to as the \textit{Baseline} model, details in Section~\ref{seqsec:evaluation}), which is similar to that in DeepLog~\cite{deeplog} and nLSALog~\cite{nlsalog}. 
%whose detail is in Section~\ref{seqsec:evaluation}.  
By applying our Baseline model upon our re-crafted key set $\mathcal{K}_1$, we find many false positive cases. We use the following FP case (session ID: -3547984959929874282) to motivate our autoencoder-based anomaly detection.

This normal session has in total 25 events, and Baseline reports the fifth event $e_5$ as an abnormal event.  The first four events are in the subsequence $s_4=[k_1, k_2, k_2, k_2]$, and the fifth event is $e_5=k_3$, where $k_1=$ \textit{``allocatedBlock ...''}, $k_2=$ \textit{``Receiving block within the localhost''}, and $k_3=$ \textit{``Received block of size 20-30 MB from 10.250.$\ast$''}.  The first ten events are listed in Table~\ref{seqtab:tnfp}.  To the Baseline model, $e_5=k_3$ is abnormal because $k_3$ is not within the top-$9\%$ predictions for $e_5$.  Top-$9\%$ predictions include variants of $k_4=$ \textit{``addStoredBlock: blockMap updated ...''}, $k_5=$ \textit{``block terminating''}, and $k_6=$ \textit{``Received block of size 60-70 MB from 10.251.$\ast$''}.  We can easily tell that both $k_3$ and $k_6$ are variants of \textit{``Received block of size $\ast$ from $\ast$''}.  In fact, the corresponding embedded vectors $\mathcal{E}(k_3)$ and $\mathcal{E}(k_6)$ are close to each other in the hyper-dimensional universe $\mathcal{U}$, 
%; furthermore, $k_3$, $k_6$, and variants of $k_4$ are in the same cluster in $\mathcal{U}$
meaning that their concepts are similar in Baseline's point of view.  
%Hence, it is understandable why Baseline thinks $k_6$ is normal, even though the size \textit{``20-30 MB''} and \textit{``60-70 MB''} are inconsistent.  

However, the fact that $k_3$ is not in top-$9\%$ but at top-$63\%$ causes this FP.  The fundamental problem is that, without the pre-knowledge of the block size-interval information for this specific session, by seeing just $s_4=[k_1, k_2, k_2, k_2]$,  Baseline would rather guess \textit{``60-70 MB''} as the block size, as it has learned through training that $k_6$ is a very frequent key (dominating $10.77\%$ of the entire dataset), whereas $k_3$ is actually an extremely rare event (only dominating $0.05\%$).  This is similar to the \textit{cold-start} problem in a recommendation system, where, not knowing personal preference, a recommendation system often recommends new users with most popular products among others.  Here, a predictor-based anomaly detection method always has to make a few bold guesses at the beginning of any sequence for the lack of information, and this issue cannot be mitigated by providing more training data (details in Section~\ref{seqsec:evaluation}). Manipulating sequence length (seqlen) cannot resolve this issue either. As long as the sequence does not include critical information, regardless of sequence length, here Baseline will always guess top predictions that may lead to FPs. It seems to us that, when knowledge is limited, Baseline is more like a detection model for extremely rare events rather than anomalies. Nevertheless, in this example, once Baseline knows the size from $e_5$, it can correctly predict the following events $e_6$, $e_7$, and $e_8$.

In contrast, an LSTM autoencoder-based anomaly detection can resolve this issue.  Unlike predictors that make guesses for next events, our autoencoder-based anomaly detection model, called \textit{DabLog}, first analyzes (encodes) the sequence, and then reconstructs (decodes) the sequence, as if the sequence is an atomic instance.  By analyzing $s_{10}=[e_1, e_2, e_3, \dots, e_{10}]$, DabLog already knows that the transmission is of size \textit{``20-30 MB''}, not \textit{``60-70 MB''}, even though $k_3$ is an extremely rare event.  In fact, DabLog could not only correctly reconstruct $s_{10}$ with $k_3$ in $e_5$'s top-$9\%$ reconstructions, but also correctly reconstruct other subsequences from $s_{11}$ to $s_{15}$, which also involve $e_5$.  Furthermore, it correctly reconstructed every subsequence from $s_{16}$ to $s_{25}$. As a result, DabLog would not falsely report this session as abnormal. In fact, with configuration $\text{seqlen}=10$ and top-$9\%$ rank-based criterion, DabLog reported 3,187 less FPs and 2,145 more TPs than Baseline upon $\mathcal{K}_1$.%; furthermore, DabLog greatly outperformed Baseline in terms of $F_1$ score upon $\mathcal{K}_0$, $\mathcal{K}_1$, and $\mathcal{K}_2$ (details in Section~\ref{seqsec:evaluation}).

\begin{table}
\caption{Example Sequential Discrete Events}
\label{seqtab:tnfp}
\centering
\begin{tabular}{ccl}
\midrule
$e_0$ & & <begin of sequence> \\
$e_1$ & $k_1$ & NameSystem.allocatedBlock /usr/root/...\\
$e_2$ & $k_2$ & Receiving block within the localhost \\
$e_3$ & $k_2$ & Receiving block within the localhost \\
$e_4$ & $k_2$ & Receiving block within the localhost \\
$e_5$ & $k_3$ & Received block of size 20-30 MB from 10.250.$\ast$ \\
$e_6$ & & blockMap updated: 10.251.$\ast$ added of size 20-30 MB \\
$e_7$ & & blockMap updated: 10.251.$\ast$ added of size 20-30 MB \\
$e_8$ & & blockMap updated: 10.250.$\ast$ added of size 20-30 MB \\
$e_9$ & & PacketResponder 1 for block terminating \\
$e_{10}$ & & Received block of size 20-30 MB from 10.251.$\ast$ \\
\bottomrule
\end{tabular}
\end{table}

\subsection{Critical Issues about Criterion}

Previous work~\cite{deeplog, nlsalog} detect anomalies by checking top-9 predictions out of 28 event keys.  Yet, this criterion is problematic because of the following reasons.
 
First, top-$9$ (top-$32\%$) is too high. By sorting the event keys by occurrence, we found that top-$9$ most frequent event keys in the dataset dominate $98.66\%$ of the entire dataset (and top-$10$ keys dominate $99.64\%$).  The coverage is so high that even a trivial model, that always blindly guess these top-$9$ event keys, can already achieve $85.58\%$ accuracy and $14.33\%$ FP rate; similarly, %a trivial model that guesses just one more predictions 
by predicting top-$10$, one can even achieve $99.16\%$ accuracy and $0.35\%$ FP rate.  Furthermore, over 10 (out of 28) event keys defined in their work~\cite{deeplog, nlsalog} did not appear in normal sequences at all; therefore, their appearance in the testing phase made their models more easier to identify anomalous sequences. Apparently, we need a much more precise model that is able to predict a smaller set of candidate events; for example, a more reasonable choice could be top-$3$ events (dominating $42.37\%$), or equivalently around top-$10\%$ of $|\mathcal{K}|$. 
Another reason for a more precise prediction (i.e., a smaller $N$ in top-$N$ prediction) is that, when we see anomalies, it would be easier to examine \textit{``what are normal''} from just top-$N$ keys to figure out \textit{``why anomalies are abnormal''}, just like our reasoning in the previous subsection.  A trivial model that guesses top-$3$ predictions would only achieve $5.71\%$ accuracy and $97.05\%$ FP rate. Finally, in the context of anomaly detection where the number of negative samples (i.e., normal events) is significantly larger then the number of positive samples, $F_1$ score is more meaningful than accuracy, because we do not care about the dominating TNs. Instead, we care more about TPs, FPs, and FNs, and hence the accuracy metric seems misleading here (details in Section~\ref{seqsec:evaluation}).

Second, the number of log keys $|\mathcal{K}|=28$ is too small.  If we look at unique sequence patterns under configuration $\text{seqlen}=10$, we have in total 28,961 patters, in which 13,056 are always normal, 11,099 are always abnormal, and 4,806 are non-deterministic. Regardless of implementation, any anomaly function $\mathcal{AD}: \mathcal{S} \rightarrow \{\text{normal}, \text{abnormal}\}$ that merely learns these 13,056 normal patterns and reports the other patterns abnormal would get reasonable results. The number of unique patterns is simply too small, and this is also the reason why these previous models needed only incredibly few training data (e.g., 4,855 normal block sessions).  However, in practice, for some applications, the number of unique event keys can easily exceed a hundred, and the patterns of normal sequences can easily become unlearnable due to the scale.  One may argue that the number of keys can be reduced by abstracting and aggregating multiple keys; however, key abstraction and aggregation may cause the loss of important information. When seeing anomalies, one may not know what what exactly happened due to lack of important information. 

%That said, we do not imply that the dataset is bad; instead, we need to re-craft the event keys so that we can get a rough idea how the previous models can work with scenarios where more event keys are involved.  We re-craft $\mathcal{K}$ into finer grained $\mathcal{K}_1$ and $\mathcal{K}_2$, where $|\mathcal{K}_1| = 101$ and $|\mathcal{K}_2| = 304$.  For $\mathcal{K}_1$, there are 220,912 normal patterns, and for $\mathcal{K}_2$ there are 1,868,327 normal patterns.  Apparently, more training data is required for the related work, and still we found that they could not provide good results even with more training data, as our re-implementation, which is trained by 200,000 normal sessions, can only achieve roughly $80\%$ $F_1$ score (details are in Section~\ref{seqsec:evaluation}). % Note that, the trivial model that blindly guess top-$30\%$ and top-$25\%$ most frequent event keys can still achieve $75.00$ and $39.35$ score for $\mathcal{K}_1$ and $\mathcal{K}_2$, respectively; apparently, we need a better approach. 
}

    {\section{Our DabLog Approach} \label{seqsec:dablog}

\subsection{Overview} 

Based on the motivation and insights in Section~\ref{seqsec:motivation}, we propose DabLog, a \textit{\textbf{D}eep \textbf{A}utoencoder-\textbf{B}ased anomaly detection method for discrete event \textbf{Log}s}. 
%We discuss a new insight: \textit{autoencoder approach may not seem as intuitive as predictors, but \textbf{autoencoder approach works and may outperform predictors} depends on application and configuration}.  
DabLog is an unsupervised and offline machine-learning model.  The fundamental differences between DabLog and the aforementioned predictor-based related work is that, DabLog determines whether $S$ is abnormal by reconstructing $S$ rather than predicting (or sometime guessing) upcoming individual events.  The intuition is that, to avoid guessing, an anomaly detection method should see a sequence as an atomic instance, and it should examine the structure of the sequence as well as the bi-directional causality among the events.  In the event of poor reconstruction, DabLog can detect \textit{not only sequences that include unseen or rare events, but also structurally abnormal sequences}.  %In the end, DabLog reports more TPs and less FPs.

DabLog focuses on \textit{\textbf{discrete events}}, which are essentially discrete-log representation derived from discrete log entries.  Each log event $e_t$  is represented as a \textit{\textbf{discrete event key}} $k_i$ (which is an abstraction string), and the key set is $\mathcal{K}=\{k_i | 1 \leq i \leq V\}$, where $V$ is the number of unique discrete events (vocabulary size). %Discrete event keys are also referred to as \textit{logkeys} in DeepLog~\cite{deeplog}, \textit{log templates} in nLSALog~\cite{nlsalog}, and \textit{discrete keys} in~\cite{lifelong}.  
%The main benefit of using discrete keys $e_t \in \mathcal{K}$ is that we do not need to understand beforehand the causality or numeric correlation among events. 
%about \textit{what are normal sequences} or \textit{what are abnormal sequences} , and hence we do not need to spend effort on repeatedly refining feature engineering.  A reduced reliance on prior knowledge means a more lightweight procedure for data preprocessing and less human interference~\cite{liuliu2}. 
Much work~\cite{kitsune22, kitsune23, liuliu3, spell} has been done for automatic discovery of unique discrete keys from security logs.  

%\textbf{\textit{Long-Term Dependency Assumption:}} We assume long-term dependencies exist among sequential system events (or operations), and sometimes the temporal spans of dependencies are long~\cite{nlsalog}.  Since in recent literature there is no clear definition of \textit{\textbf{how long}} a long-term dependency should be, we consider any \textit{\textbf{non-immediate}} dependency as long-term dependency.  A non-immediate dependency exists between two events $e_i$ and $e_j$ if $e_i$ and $e_j$ has causality (or correlation) but there always is at least one $e_k$ (where $i<k<j$) sitting in between $e_i$ and $e_j$; for example, $e_i$ and $e_j$ can be \textit{``receiving an object from somewhere''} and \textit{``received an object''}, and if it takes time to transfer such an object then very likely there are many other events sitting in between $e_i$ and $e_j$.  As opposed to non-immediate dependencies, immediate dependencies exist between $e_i$ and $e_j$ such that $j-i \leq \epsilon$; for example, $e_i$ and $e_j$ can be \textit{``ask to delete a remote object''} and \textit{``deleting a remote object''}, and if the approval is always immediately granted then very likely $j-i=1 \leq \epsilon$ (e.g., $1 \leq \epsilon \leq 3$).

We make a \textbf{\textit{Time-Sensitive Distribution Assumption: }} we assume that the value of the time-sensitive distribution $\mathcal{D}$ of an event $e_t$ at time $t$ may depend on both the past and future events; that is, one can expect both past and future causal events $e_i$ and $e_j$ when observing an event $e_t$, where $i<t<j$.  For example, if $e_t$ is \textit{``deleting a remote object''}, then one can expect there is a past event $e_i$ like \textit{``ask to delete a remote object''} and a future event $e_j$ like either \textit{``deleted an remote object''} or \textit{``deletion error''} (if such audit logs are available).  We define the probability function of $e_t$ for $t$ in $i \leq t \leq j$ by $\text{Pr} (e_t | e_i, e_{i+1}, \dots, e_j)$; this assumption is not applicable to predictors, whose probability mass functions are typically defined by only the past, that is $\text{Pr} (x_t |x_{t-1}, \dots, x_1)$.

In summary, DabLog aims to provide an anomaly detection function $\mathcal{AD}: \mathcal{S} \rightarrow \{\text{normal}, \text{abnormal}\}$.  DabLog consists of four major components (Figure~\ref{seqfig:model.pdf}): an embedding layer, a deep LSTM autoencoder, an event classifier, and an anomaly critic.  The workflow is stated as follows.  Given a sequence $S \in \mathcal{S}$, the embedding layer $\mathcal{E}$ embeds $S$ into an embedded distribution $X_e$, the autoencoder then analyzes (encodes) $X_e$ and reconstructs (decodes) the categorical logit distribution $Y$, the event classifier then transforms $Y$ into categorical probability distribution $\mathcal{P}$, and lastly the critic compares $\mathcal{P}$ with $S$ and reports whether $S$ is normal or abnormal.

\seqfig{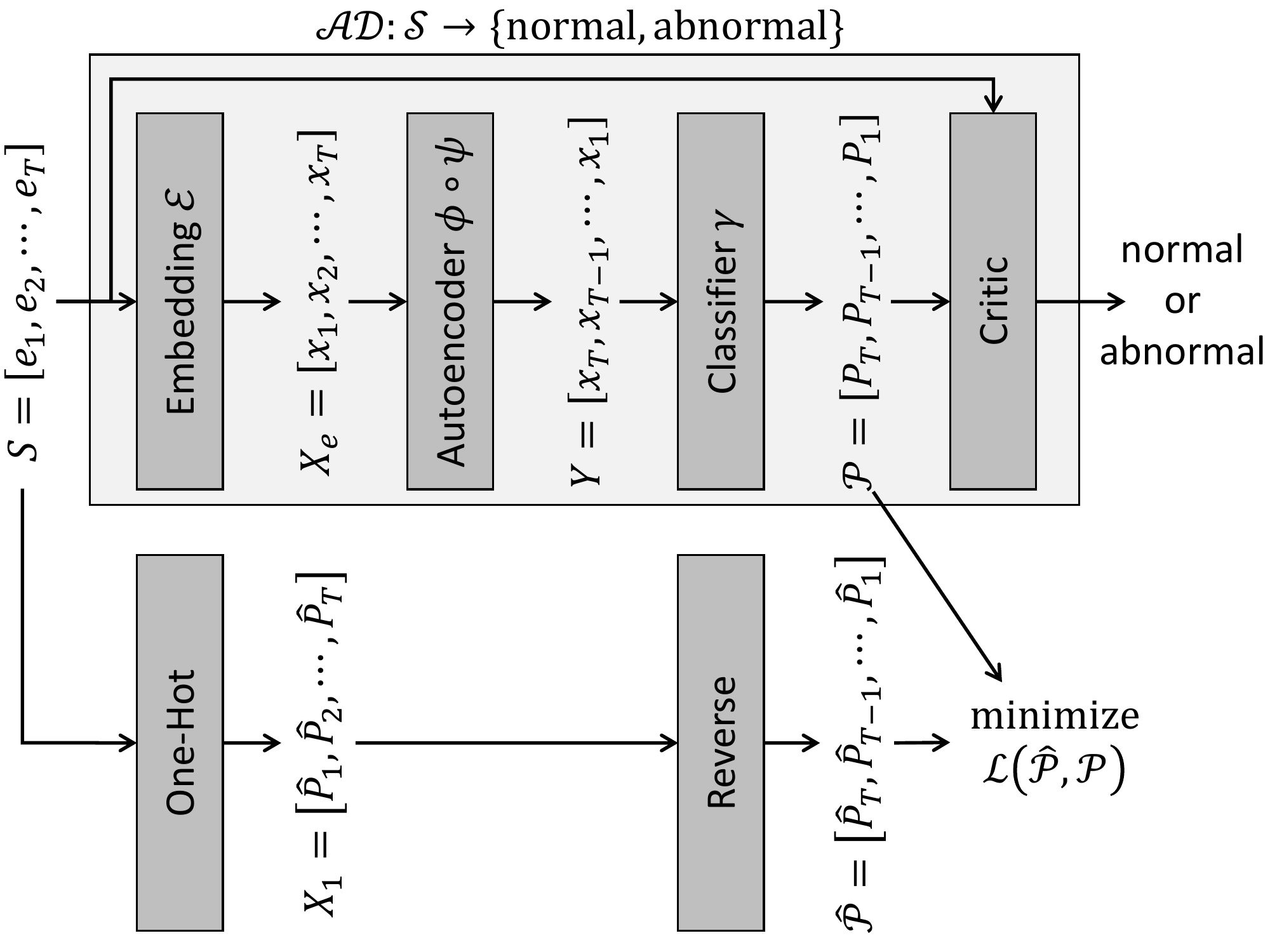}{}{DabLog Anomaly Detection Model}

\subsection{Embedding Layer}

Since our anomaly detection takes a sequence of discrete events $S=[e_t | 1 \leq t \leq T]$ as input, we need to embed discrete events $e_t \in \mathcal{K}$ into a particular model-recognizable vector, where $\mathcal{K} = \{k_i | 1 \leq k \leq V\}$ is the set of discrete event keys of vocabulary size $V=|\mathcal{K}|$.  We denote an embedding function as $\mathcal{E}: \mathcal{S} \rightarrow \mathcal{X}$ and the procedure as $X_e=\mathcal{E}(S)$, where $X_e=[x_t | 1 \leq t \leq T ] \in \mathcal{X}$ is an embedded distribution of $S$, and $x_t$ is the embedded vector of $e_t$.  There are three common embedding options adopted by prior work: (1) embedding with one-hot representation, (2) embedding using pre-trained natural linguistic packages, and (3) embedding by training an additional embedding layer along with the other layers.  

We adopt the last option, because the other two options have major drawbacks.  
On one hand, one-hot representation, in which $x_t=[v_i | 1 \leq i \leq V]$ where $v_i\in\{0, 1\}$ and $\sum_i v_i = 1$, not only lacks the ability to embed the semantic or correlation features among keys, but also causes the model to suffer from dimension explosion when $V$ is large (dimension explosion causes run-time inefficiency in machine learning). 
As a consequence, leveraging one-hot representation, DeepLog~\cite{deeplog} does not work well on datasets with more keys (even the HDFS dataset), even though its accuracy has been improved by stacking two LSTM layers.
On the other hand, while directly using pre-trained natural linguistic packages (e.g., Word2Vec and GloVe) seems convenient, it may not work well on security audit logs that lack natural linguistic properties~\cite{liuliu2}.  The reasons include that (1) a JSON-formatted or CSV-formatted log may not demonstrate syntactic structure, (2) duplicate or redundant attributes may introduce unwanted noise, and (3) arbitrary abbreviated strings may not have a match in the existing packages.  

Although the last option \textit{training an additional embedding layer} is slower, the embedding function $\mathcal{E}$ can be well customized for the specific log dataset.  That is, rather than no correlation (with one-hot representation) or syntactic correlation (using linguistic packages), the underlying correlation between discrete events $k_i \in \mathcal{K}$ (for example, $k_3$ and $k_6$ in Table~\ref{seqtab:tnfp}) can be found by $\mathcal{E}$.  

In our approach, an embedding layer is instantiated by $V$ and the size of output dimension $\delta$, and then it holds a random matrix that maps $e_t=k_i$ to $x_t$, where $x_t$ is an embedded vector of size $\delta$.  This matrix is then trained by back propagation during its training phase along with the time-sensitive encoder-decoder network.  
In addition to event keys, we incorporate three special padding keys \textit{begin-of-sequence}, \textit{end-of-sequence} and \textit{unknown} in our embedding layer.  
On one hand, the keys \textit{begin-of-sequence} and \textit{end-of-sequence} provide additional sequential characteristics to LSTM models, and we noticed a slight improvement for both autoencoders and predictors in detection results.  On the other hand, the \textit{unknown} key is used for improving computational performance.  The problem of not using \textit{unknown} key is that, the embedding function in prior work initiates untrained embedding vectors for all unknown events, and similarly the event classifier also initiates unused logit dimensions for unknown events.  It is inefficient to train such a machine-learning model when unknown events unnecessarily use much resource.   

\subsection{Deep LSTM Autoencoder}

Deep autoencoders have been used in time-insensitive anomaly detection.  Conceptually, an autoencoder learns the identity function of the normal data and reconstructs normal data distribution; hence the input data leading to poor reconstruction is potentially abnormal.  Since we are tackling time-sensitive discrete events instead of engineered features, our autoencoder is different from typical ones that reconstruct the input features.  Rather, it tries to reconstruct the logit distribution of categorical events.

A typical autoencoder is trained by minimizing the function:
$ \phi, \psi = \text{arg min}_{\phi, \psi} \|X - (\psi \circ \phi) (X) \| $, where $\phi$ is an encoder, $\psi$ is a decoder, $X$ is the input distribution, and $\psi \circ \phi (X)$ is the target (reconstructed) distribution.  To tackle time-sensitive discrete events, our autoencoder (Figure~\ref{seqfig:autoencoder.pdf}) is trained by minimizing the function:
\begin{align*}
\phi, \psi & = \underset{\phi, \psi}{\text{arg min}} \| X - Y \| ^ 2 \\
& = \underset{\phi, \psi}{\text{arg min}} \| \text{rev}(X_e) - (\psi \circ \phi) (X_e) \| ^ 2 \\
& = \underset{\phi, \psi}{\text{arg min}} \|(\text{rev} \circ \mathcal{E})(S) - (\psi \circ \phi \circ \mathcal{E})(S) \| ^ 2 
\end{align*}
where $\phi$ is a deep encoder, $\psi$ is a deep decoder, and $\text{rev}$ is a function that reverses a distribution matrix.  The encoder $\phi$ maps an $\mathcal{E}$-embedded matrix $X_e = \mathcal{E}(S)$ into a representation $\text{code}=\phi(X_e)$, whereas the decoder $\psi$ maps the \text{code} into a target distribution matrix $Y= \psi(\text{code})$.  Hence, the reconstructed distribution through the embed-encode-decode procedure is denoted as $Y=(\psi \circ \phi \circ \mathcal{E}) (S)$.  The function \text{rev} is involved because $Y$ is in the reverse order from $X_e$ due to LSTM's hidden state $h_t$, which is explained below.

We build our encoder $\phi$ and decoder $\psi$ by stacking vanilla Long Short-Term Memory (LSTM)~\cite{lstm} (variants are applicable as well~\cite{lstmvariants1, lstmvariants2}).  The advantage of using an recurrent LSTM network over traditional recurrent neural networks is that an LSTM unit calculates a hidden state that conceptually remembers past activities as well as long-term dependencies~\cite{longtermisdifficult}.  For presentation purpose, we denote the computation of hidden state $h_t$ by 
%$h_t = H (x_t, h_{t-1})$.  
\begin{align*}
h_t = \text{LSTM } (x_t, h_{t-1})    
\end{align*}
That is, at each time-step $t$, an LSTM unit takes two inputs $x_t$ (the current data point) and $h_{t-1}$ (previous hidden state), and it generates an output $h_t$ (current hidden state).  Multiple LSTM layers each calculates its own hidden state, as illustrated in Figure~\ref{seqfig:autoencoder.pdf}.  The representation $\text{code}=\phi (\mathcal{E}(S))$ is essentially the transferable hidden state $h_T$ at the last time-step $T$, and $h_T$ is calculated by applying LSTM function iteratively from $t=1$ up until the last time-step $t=T$.  
%LSTM iteratively encodes $x_t$ into the representation code $h_T$, and w
We can conceptually think of this procedure as \textit{``pushing $x_t$ into a state stack $h_T$''}; hence, the conceptual procedure for decoder is \textit{``popping $x_t$ out from a state stack $h_T$''}.  Therefore, the distribution $X_e$ and $Y$ are in reverse order.

%From Equation~\ref{seq:design:hiddenstate}, we can expand-and-abbreviate $h_t$ to as (assuming $W_\ast$ and $b_\ast$ are already trained):
%\begin{align*}
%&& h_t = & \text{ } o_t \times \text{tanh}(C_t) \\
%&& = & \text{ } o_t \times (f_t \times C_{t-1} + i_t \times \tilde{C}_t) \\
%&& = & \text{ } O (h_{t-1}, x_t) \times (F (h_{t-1}, x_t) \times C_{t-1} \\
%&&  & \text{ } + I (h_{t-1}, x_t) \times \tilde{\mathcal{C}} (h_{t-1}, x_t)) \\
%&& = & \text{ } H (h_{t-1}, x_t, C_{t-1})
%\end{align*}
%where $O, F, I, \tilde{\mathcal{C}}$ and $H$ are functions for deriving $o_t, f_t, i_t, \tilde{C}_t$ and $h_t$ (from $h_{t-1}, x_t$ and $C_{t-1}$), respectively.  

Our deep LSTM autoencoders are similar to traditional autoencoders that consist of deep encoders and deep decoders, except that our encoder $\phi$ and decoder $\psi$ are implemented with stacked LSTMs (of at least two layers).  The number of hidden units decreases (e.g., by half) layer-by-layer in $\phi$, and increases (e.g., doubled) layer-by-layer in $\psi$.  Some research work~\cite{deeprnn1, deeprnn2, deeprnn3} have addressed the main benefit of stacking multiple LSTM layers over using a single layer: stacking hidden states potentially allows hidden states at each layer to reflect information at different timescale, and the final layer can gain benefits from learning or combining representations given by prior layers (hence better results).

Our autoencoder is \textit{unconditional}, meaning that we do not provide a condition $\hat{y}_{\tau}=e_{T-\tau+1}$ to the decoder $\psi$ when it is decoding $y_{\tau+1}$ for any $y_{\tau}$ in $Y=[y_\tau|1 \leq \tau \leq T]$.  This decision is made differently from some predictor-based anomaly detection methods~\cite{deeplog, habad, nlsalog, nlsalog9, nlsalog12} that either provide $e_{T+k-1}$ to $\psi$ when decoding $e_{T+k}$ or predict only $e_{T+1}$ for any input sequence $S=[e_t | 1 \leq t \leq T]$.  Srivastava et al.~\cite{lstmae2} have shown that, although conditional decoders could provide slightly better results in predictors, unconditional decoders are more suitable for autoencoders.  There are two reasons.  First, autoencoders have only one expected output from any input sequence, which is the reconstruction of the input, whereas predictors could have multiple expected outputs (say $S_1$ and $S_2$ have the same prefix of length $T$ but different suffices). While the condition acts as a hint about which suffix should be decoded in predictors, providing conditions serves no additional purpose in autoencoders.  Second, usually there is strong short-term dependency among adjacent events, and hence it is not ideal to provide a condition that may cause the model to easily pick up short-term dependencies but omit long-term dependencies.

\subsection{Event Classifier and Anomaly Critic}

Since our goal is to provide an anomaly detection method $\mathcal{AD}: \mathcal{S} \rightarrow \{\text{normal}, \text{abnormal}\}$, simply combining an embedding layer and a deep LSTM autoencoder will not accomplish our goal.  Similar to prior predictor-based anomaly detection methods~\cite{deeplog, nlsalog, nlsalog9, nlsalog12}, right after our deep autoencoder, we add an additional single-layer fully connected feed-forward network $\gamma$, which is activated by a \textit{softmax} function.  The last layer $\gamma$ acts as a multi-class classifier that takes input $Y$ (which is the reconstructed distribution from $\psi$) and generates a probabilistic matrix $\mathcal{P}=[ P_\tau | 1 \leq \tau \leq T]$, where $P_\tau=[ p_i | 1 \leq i \leq V ]$ and $p_i$ can be interpreted as the likelihood of the discrete event $e_t=e_{T-\tau+1}$ being an instance of discrete event key $k_i$ (that is, $\gamma$ is an event classifier).  We explain why we need $\gamma$ in the following paragraph.  In order to train $\gamma$, one-hot representation of $S$ is provided as true probabilistic matrix, denoted as $X_1=\text{onehot}(S)=[\hat{P}_t | 1 \leq t \leq T]$, where $\hat{P}_t=[\hat{p}_i | 1 \leq i \leq V]$ and $\hat{p}_i \in [0, 1]$ and $\sum_i \hat{p}_i=1$.  Similar to the embedding function $\mathcal{E}$, we also include three additional special padding keys \textit{begin-of-sequence}, \textit{end-of-sequence}, and \textit{unknown} in the onehot function.  Note that $X_1$ and $\mathcal{P}$ are in reverse order, so $\hat{\mathcal{P}} = \text{rev} (X_1)$.  In our design, the multi-class classifier $\gamma$ is trained by minimizing the categorical cross-entropy loss function:  
\begin{align*}
\mathcal{L} (\hat{\mathcal{P}}, \mathcal{P}) & = \sum_\tau^T L (x_{T-\tau+1}, P_\tau) \text{, where} \\
L (x_t, P_\tau) & = - \sum_i^V \hat{p}_i \times \text{log} (p_i) \\
\end{align*}

In summary, the overall embedder-encoder-decoder-classifier network tries to minimize the function:
\begin{align*}
\mathcal{E}, \phi, \psi, \gamma & = \underset{\mathcal{E}, \phi, \psi, \gamma}{\text{arg min}} \| (\text{rev} \circ \text{onehot})(S) - (\gamma \circ \psi \circ \phi \circ \mathcal{E})(S) \|
\end{align*}

Unlike typical time-insensitive autoencoder-based anomaly detection methods, we do not directly use scalar reconstruction errors (e.g., root-mean-square error) as anomaly scores.  The reason is that our problem---\textit{identifying time-sensitive anomaly by examining discrete events}---is more like a language processing problem.  We can view sequences $S$ as sentences and events $e_t$ as words, and we care more about wording (e.g., \textit{``which words $e_t$ better fit in the current sentence $S$''}) rather than embedding (e.g., \textit{``which vector $y_\tau$ better vectorize $e_t$ in the current sentence $S$''}).  As such, we need the event classifier $\gamma$ as well as certain reasonable \textit{``wording options''} that help us with finding \textit{``fitting words''} and \textit{``unfitting words''}, or equivalently, normal events and abnormal events in $S$.

Rank-based criterion and threshold-based criterion are two common \textit{``wording options''} adopted by previous predictor-based anomaly detection methods.  Say a discrete event $e_t$ is an instance of $\hat{k}_i$, a rank-based criterion will consider $e_t$ anomalous if $p_i$ is not in top-$N$ prediction (e.g., $N=V/10$) in $P_\tau$, and a threshold-based criterion will consider $e_t$ anomalous if $p_i \in P_\tau$ is under a particular threshold $\theta_P$.  In other words, discrete keys $\{k_j | \forall j \text{ s.t. } p_j \in \text{arg top}_N (P_\tau)\}$ and $\{k_j | \forall j \text{ s.t. } p_j > \theta_P \}$ are normal discrete keys.  Our autoencoder-based anomaly detection adopts both threshold-based and rank-based criteria, but for presentation purpose we demonstrate the rank-based criterion (Figure~\ref{seqfig:topn.pdf}) in this paper in order to compare our work with prior predictor-based methods~\cite{deeplog, nlsalog}.  However, both rank-based and threshold-based criteria have the same major drawback: they brutally divide $P_\tau$ while omitting the correlation between the true $\hat{k}_i$ and the supposedly normal keys; hence, besides the aforementioned two criteria, our anomaly detection has an option of a novel criterion, which is discussed in Section~\ref{seq:discussion:criterion}.

Similar to prior anomaly detection work~\cite{deeplog, nlsalog} that conducted experiments on the same dataset~\cite{sosp2009, icdm2009}, in which anomaly labels (e.g., normal or abnormal) are given at the sequence level, our anomaly detection model gives labels to sequences.  We say a sequence $S=[e_t | 1 \leq t \leq T]$ is abnormal if any $e_t \in S$ is abnormal.  With aforementioned criteria, our anomaly detection method $\mathcal{AD}: \mathcal{S} \rightarrow \{\text{normal}, \text{abnormal}\}$ is complete.

\seqfig{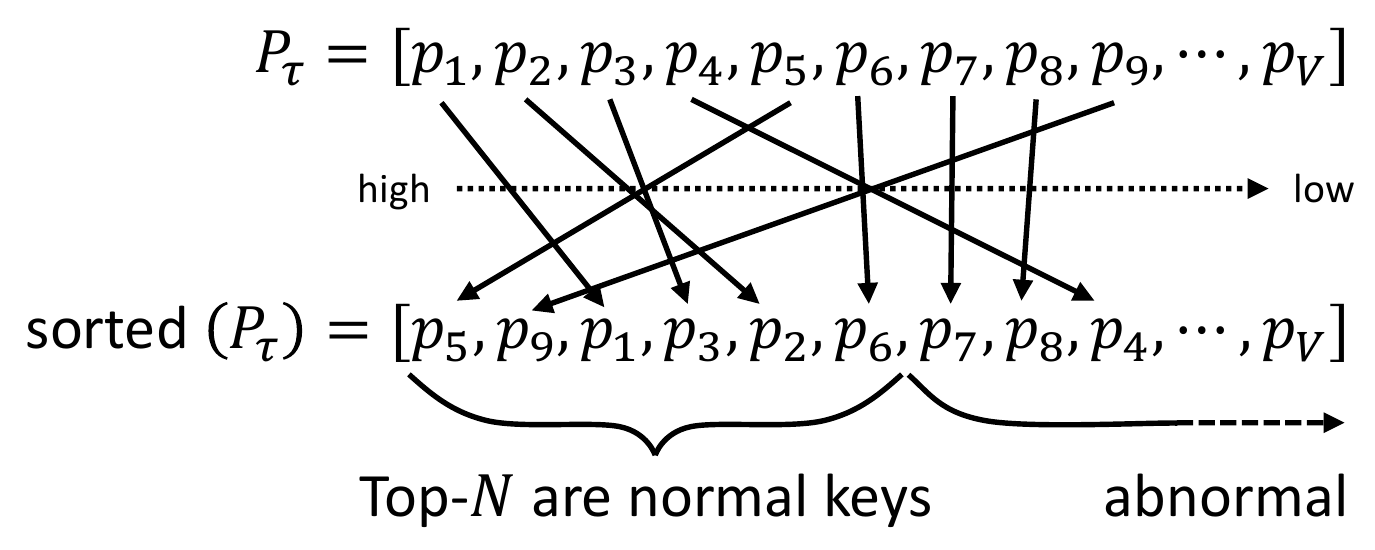}{0.9}{An example of the rank-based criterion}

}

    {\section{Evaluation} \label{seqsec:evaluation}

We motivate our evaluation with three questions:
(1) how better is DabLog in comparison with a predictor-based baseline model,
(2) how does \textit{having more keys} impact the detection results,
and (3) how does the fundamental difference make DabLog more advantageous.
Similar to prior work~\cite{deeplog,nlsalog}, we evaluate DabLog with the Hadoop File System (HDFS) console-log dataset released by Xu et al.~\cite{sosp2009}.  We show that DabLog not only is capable of detecting system anomalies, but also outperforms our predictor-based re-implementation baseline model.

\subsection{DabLog and Baseline Implementation}

We implement both Baseline and DabLog models with 3K lines of Python 3.7.4 script, and we leverage deep-learning utilities from package Tensorflow 2.0.0.  

\textit{\textbf{Baseline Model Implementation:}} We re-implement an anomaly detection model which we believe is representative of predictor-based models (Figure~\ref{seqfig:baseline.pdf}).  Similar to DeepLog~\cite{deeplog} and nLSALog~\cite{nlsalog}, the \textit{Baseline} model has a two-layered LSTM network, a multi-class classifier, and a rank-based critic, except that unlike DeepLog it does not learn one-hot representation, and unlike nLSALog it does not include a self-attention layer. The embedding layer is implemented with a \textit{tensorflow.keras.layers.Embedding} layer, in which we also include three additional special padding keys \textit{begin-of-sequence}, \textit{end-of-sequence}, and \textit{unknown} (we notice a slightly improvement for both models when including them). The two-layered LSTM network is implemented by stacking two \textit{tensorflow.keras.layers.LSTM} layers, and each is activated by ReLU.  Each LSTM layer is configured to have 64 hidden units just like the ones in DeepLog and nLSALog.  Lastly, the event classifier layer is implemented with \textit{tensorflow.keras.layers.Dense}, which is activated by the \textit{softmax} function, just like the ones in DeepLog and nLSALog.  In event classifier, we also include the three additional special padding keys.

Note that we do not reuse DeepLog's code for the following reason. The only difference between Baseline and DeepLog is that DeepLog uses one-hot representation, whereas Baseline (as well as nLSALog) uses an embedding layer.  Since the performance difference has already been addressed in prior predictor-based work (e.g.,~\cite{nlsalog, nlsalog9, habad, nlsalog12}) and in the research field of natural-language processing (e.g.,~\cite{wordembedding}), we believe it is redundant to re-evaluate the original DeepLog implementation.  Still, the other parameters (including the numbers of layers and the numbers of hidden units) used in Baseline are the same as in DeepLog and nLSALog.

\textit{\textbf{DabLog Model Implementation:}} Our encoder and decoder are implemented by stacking two \textit{tensorflow.keras.layers.LSTM} layers, and each is activated by ReLU.  The encoder is configured to have 64 and 32 hidden units for its 1\textsuperscript{st} and the 2\textsuperscript{nd} layer respectively, and the decoder is configured to have 32 and 64 hidden units for its 1\textsuperscript{st} and the 2\textsuperscript{nd} layer respectively, as we follow the common practice that the representation code is a downgraded abstraction.  The embedding layer and the event classifier are implemented in the same way as the ones in the \textit{Baseline} model.  In DabLog, the encoder network is connected with the decoder by \textit{tensorflow.keras.layers.RepeatVector}, and the decoder network is connected with the classifier by \textit{tensorflow.keras.layers.TimeDistributed}.

To train the DabLog and Baseline, Adam optimizer is used with accuracy metric in minimizing categorical cross-entropy.  We use the same sequence-length configuration $\text{seqlen}=10$.  Sliding window is applied to longer sequences $S$ that have lengths $|S| > \text{seqlen}$.  %For Baseline, the slideing window of $S$ are $\{s_T | 1 \leq T < |S|$, where $s_T=[e_t | \text{max} (1, T-9) \leq t \leq T]$.  For DabLog, the sliding windows are $\{s_T | 10 \leq T \leq |S| \}$, where $s_T=[e_t | T - seqlen < t \leq T]$.  

\subsection{Experiment Setup}

The HDFS dataset~\cite{sosp2009} encloses over 11 million log entries from Hadoop map-reduce jobs that ran on 203 Amazon EC2 nodes across two days.  Each log entry contains a block identifier, and each block can be understood as a concurrent thread (i.e., log entries that have the same block identifier are executed sequentially).  The anomaly labels (i.e., normal or abnormal) are provided at the block level, and there are 558,223 normal blocks and 16,838 abnormal blocks.  Xu et al.~\cite{icdm2009} addressed that the labels were given based on 680 unique event traces across all the data, and an event trace is labeled as normal if it contains all the events of a given pattern.  

\subsubsection{\textbf{Dataset Engineering}}

As mentioned in Section~\ref{seqsec:motivation}, we think the number of event keys $|\mathcal{K}|=28$ is too small.  To measure how well the related work can be applied to scenarios where more event keys are involved (note that we do not re-craft keys for performance improvement---having more keys surely reduces the prediction performance), we re-crafted the event keys into three sets $\mathcal{K}_0$ (new base), $\mathcal{K}_1$, and $\mathcal{K}_2$, so that we have 31, 101, and 304 keys, respectively.  Since anomaly labels are given upon sequences (by block ID), it is safe to transform keys without modifying the original sequences (that is, anomaly labels are not impacted by our key transformation).  The statistics of each key set under configuration $\text{seqlen}=10$ is listed in Table~\ref{seqtab:logkeys10}.  These key sets are from the same source log, except that $\mathcal{K}_1$ and $\mathcal{K}_2$ discard less information by re-attaching add-on strings; for example:
\begin{align*}
&& k_i \in \mathcal{K}_0 : & \textit{``Received block''} \\
&& \text{1\textsuperscript{st} add-on} : & \textit{``of size 20-30 MB''} \\
&& \text{2\textsuperscript{st} add-on} : & \textit{``from 10.250.$\ast$''} \\
&& k_j \in \mathcal{K}_1 : & \textit{``Received block of size 20-30 MB from 10.250.$\ast$''} 
\end{align*}
There are three types of add-on strings.  First, for two event keys that each involves a filepath, we attach one of the 32 filepath add-ons (e.g., \textit{``/user/root/randtxt/\_temporary/\_task\_$\ast$/part$\ast$''} and \textit{``/mnt/hadoop/mapred/system/job\_$\ast$/job.jar''}) we got from manual analysis.  Second, for two event keys that each involves a filesize, we attach one of the seven 10-MB interval add-ons (e.g., \textit{``0-10 MB''} and \textit{``60-70 MB''}).  Third, for nine events that each involves an IP address, we attach add-ons from the following rules.  

\begin{enumerate}
\item If an event key involves both a source IP address and a destination IP address, we check and attach add-ons that represent whether it is \textit{``within the localhost''}; if not, we then check whether it is \textit{``within the subnet''} or \textit{``between subnets''} by the IP prefixes (e.g., 10.251.7$\ast$)
\item If an event key involves either a source IP address or a destination IP, we attach add-ons that represent directions and IP prefixes (e.g., \textit{``from 10.251.7$\ast$''}).
\end{enumerate}

The IP-prefix granularity is different for $\mathcal{K}_1$ and $\mathcal{K}_2$: for $\mathcal{K}_1$ we use the first two decimal numbers (e.g., 10.251.$\ast$), and for $\mathcal{K}_2$ we use just one more decimal number (e.g., 10.251.7$\ast$).  We split $\mathcal{K}_0$ by attaching add-ons, but we also discard keys that have zero occurrence.  
%Among these key sets, we think $\mathcal{K}_1$ is the most representative.  Note that in the context of anomaly detection, a model learns only from a subset of the normal dataset (e.g., partial normal patterns and partial undecidable patterns) in hope of identifying abnormal patterns.  While our method of splitting event keys does not look perfect, we argue that, in the anomaly detection literature, the point of applying machine learning to distribution of discrete events rather than engineered features is that processing discrete events does not require much domain knowledge or heavy human efforts.  In this evaluation, we do not intent to make DabLog aware of any domain details, including whether the event pre-processing is perfect or not.  As such, we think our processing method is reasonable for evaluation purpose.
Among these key sets, we think $\mathcal{K}_1$ is the most representative.  Note that one benefit of applying machine learning to discrete events is that, detail domain knowledge is no longer required.  While our method of splitting event keys does not look perfect, we argue that, this method serves our purpose of evaluating the performance of DabLog and Baseline at different number of logkeys, as we do not intent to make them know any domain details.

\begin{table}
\caption{Statistics of Event Key Sets under $\text{seqlen}=10$}
\label{seqtab:logkeys10}
\centering
\begin{tabular}{rrrrr}
\toprule
& $|\mathcal{K}_*|$ & Normal & Abnormal & Undecidable \\
& Size & Patterns & Patterns & Patterns \\
\midrule
$\mathcal{K}_0$ & 31  & $13,056$    & $11,099$ & $4,806$   \\ 
$\mathcal{K}_1$ & 101 & $220,912$   & $35,662$ & $19,925$  \\
$\mathcal{K}_2$ & 304 & $1,868,327$ & $103,863$ & $49,856$ \\
\bottomrule
\end{tabular}
\end{table}

\seqfig{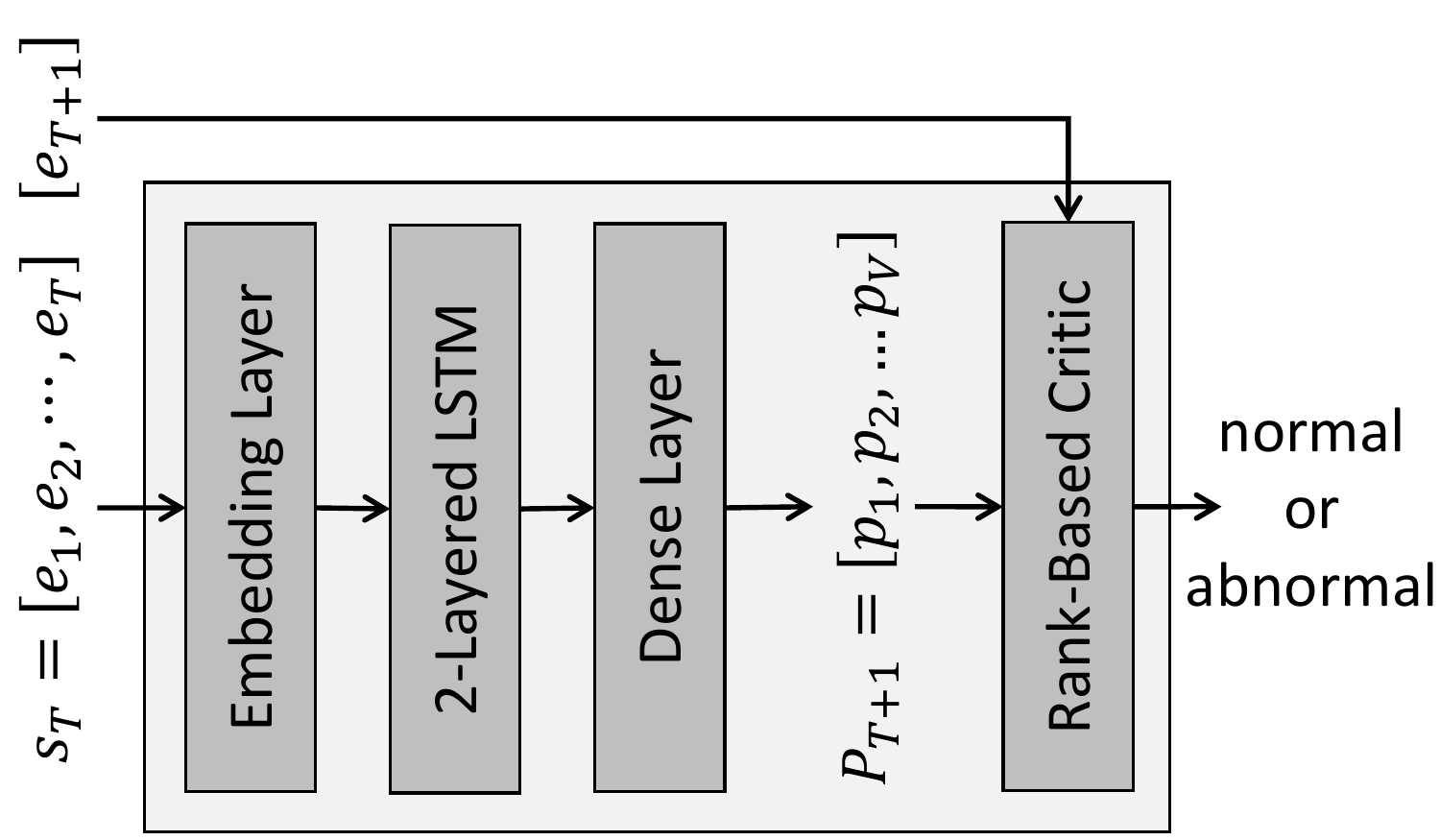}{0.8}{Baseline Predictor-Based Model}

\subsubsection{\textbf{Training Datasets and Testing Datasets}}

In the context of anomaly detection, a model learns whatever it needs from partial normal dataset, and uses its knowledge to identify anomalies.  DeepLog and nLSALog suggested that, upon $|\mathcal{K}|=28$, a training dataset which included only 4,855 normal sessions was large enough.  Similarly, we train DabLog and the baseline model with 5,000 normal sessions, and we denote the resulting model as \textit{DabLog (5K)} and \textit{Baseline (5K)}, respectively.  However, since these 5K sessions cannot cover the majority of the normal patterns in experiments upon $\mathcal{K}_1$ and $\mathcal{K}_2$, we also train DabLog and the baseline model with larger datasets, and we denote the resulting model as \textit{DabLog} and \textit{Baseline}.  The training dataset for $\mathcal{K}_0$ and $\mathcal{K}_1$ consists of 200,000 normal sessions, and the training dataset for $\mathcal{K}_2$ consists of 100,000 normal sessions.  The number of sessions in dataset for $\mathcal{K}_2$ is smaller due to the computational limitation within our experiment environment.  The testing datasets for $\mathcal{K}_0$, $\mathcal{K}_1$, and $\mathcal{K}_2$ each includes all 16,868 abnormal sessions.  Beside abnormal sessions, the testing dataset for $\mathcal{K}_0$ and $\mathcal{K}_1$ include 200,000 normal sessions, and the testing dataset for $\mathcal{K}_2$ include 100,000 normal sessions.

\begin{figure*}
\centering
\subfigure[$F_1$ Scores upon $|\mathcal{K}_0|=31$ Event Keys]{
\includegraphics[width=0.32\textwidth]{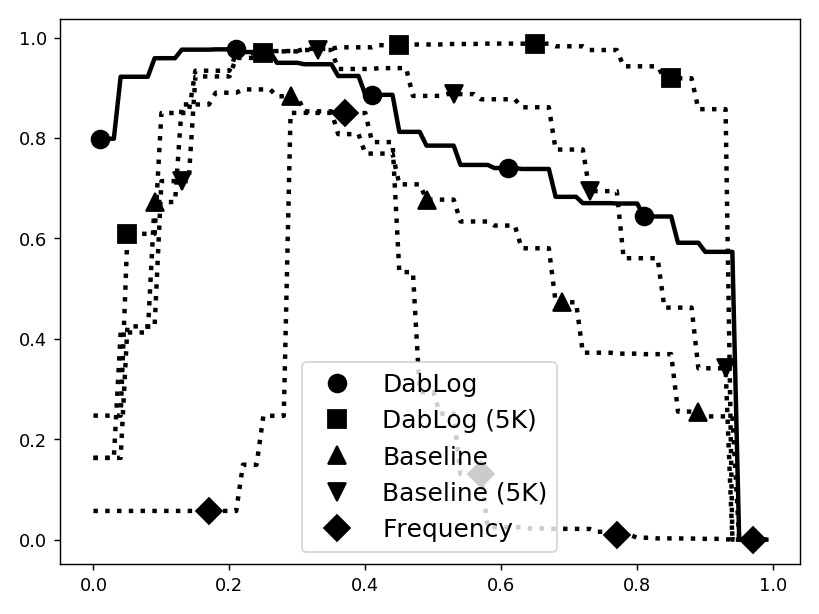}\label{seqfig:logkeys31.png}} 
\subfigure[$F_1$ Scores upon $|\mathcal{K}_1|=101$ Event Keys]{
\includegraphics[width=0.32\textwidth]{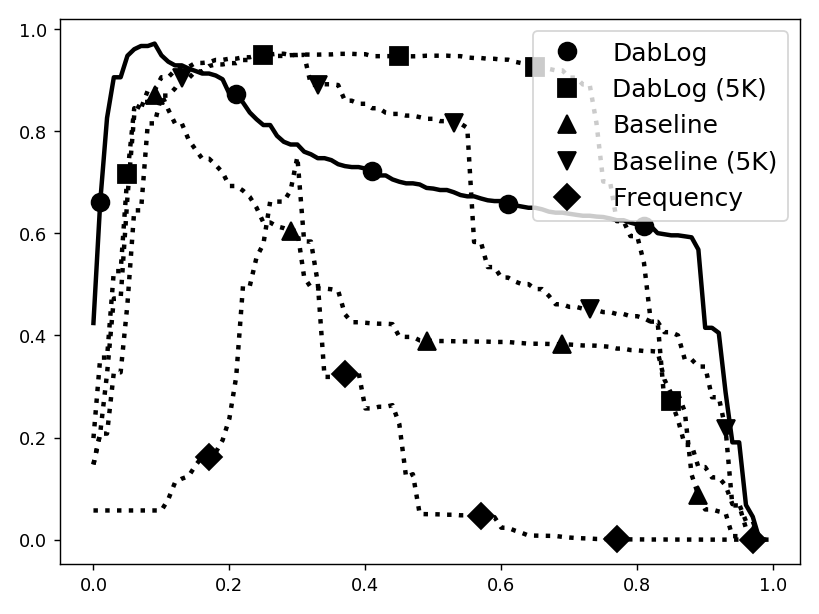}\label{seqfig:logkeys101.png}}
\subfigure[$F_1$ Scores upon $|\mathcal{K}_2|=304$ Event Keys]{
\includegraphics[width=0.32\textwidth]{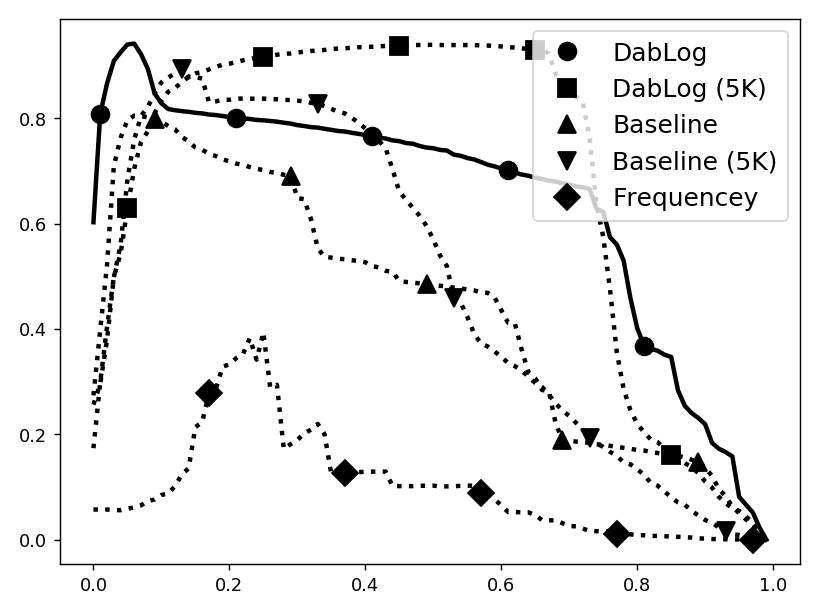}\label{seqfig:logkeys304}} 
\subfigure[DabLog's Metrics upon $|\mathcal{K}_1|=101$ Event Keys]{
\includegraphics[width=0.32\textwidth]{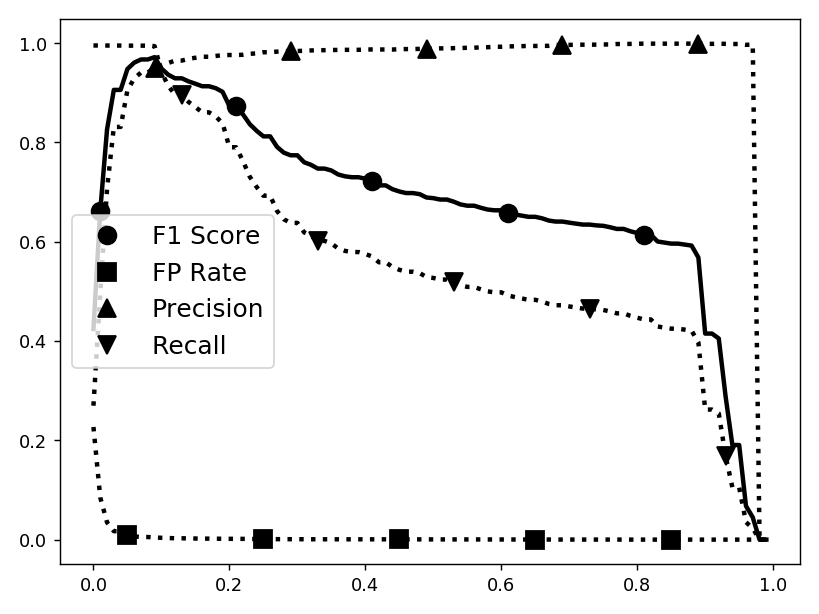}\label{seqfig:dablog101.png}}
%\subfigure[DabLog's Metrics upon $|\mathcal{K}_2|=304$ Event Keys]{
%\includegraphics[width=0.32\textwidth]{seqfigures/dablog304.png}\label{seqfig:dablog304.png}}
\subfigure[Precision upon Different Key Sets]{
\includegraphics[width=0.32\textwidth]{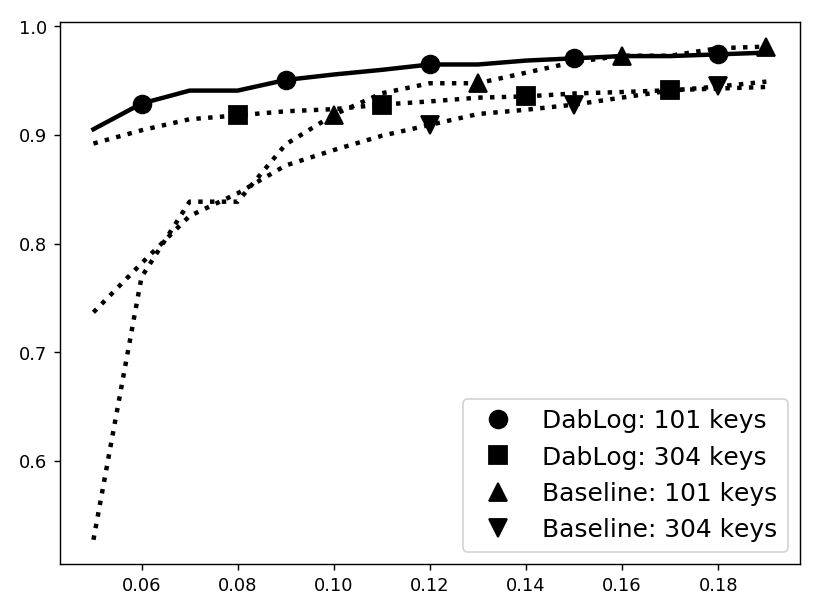}\label{seqfig:precision.png}} 
\subfigure[Recall (TP Rate) upon Different Key Sets]{
\includegraphics[width=0.32\textwidth]{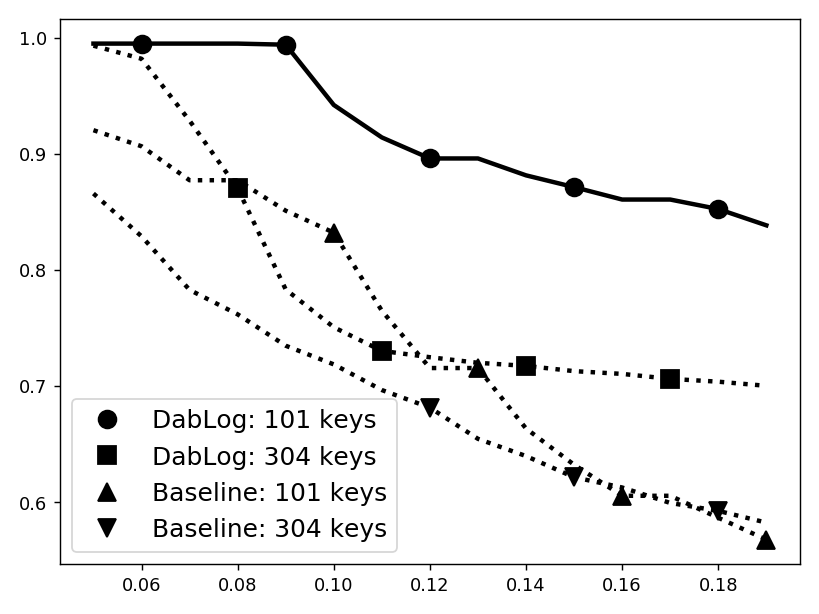}\label{seqfig:recall.png}} 
\caption{Metrics Comparison Among Different Models and Different Configurations (System-Log Dataset)}
\label{seqfig:figureset1}
\end{figure*}

\subsection{Anomaly Detection Results}\label{seqsec:evaluation-f1}

To compare different models, we leverage the $F_1$ score metric, whose equations are listed below, where $TP$, $FP$, $TN$, and $FN$ denote the numbers of true positives, false positives, true negatives, and false negatives, respectively.  The higher the $F_1$ score, the better the model in providing good anomaly detection results.  Previous work was evaluated by the accuracy metric.  However, we believe the accuracy metric is misleading for imbalanced dataset, as a blind model that always returns normal for any sequences can achieve high accuracy due to the fact that $TN \gg FN$. 

\begin{align*}
\text{Recall} & = \text{TP Rate} = \frac{TP}{TP+FN} &
\text{Precision} & = \frac{TP}{TP+FP} \\
F_1 \text{ Score} & = 2 \times \frac{\text{Precision} \times  \text{Recall}}{\text{Precision} + \text{Recall}} &
\text{FP Rate} & = \frac{FP}{FP+TN} \\
\text{Accuracy} & = \frac{TP+TN}{TP+TN+FP+FN}
\end{align*}

Figure~\ref{seqfig:logkeys31.png} depicts the $F_1$ score trends of different models upon the base key set $\mathcal{K}_0$.  The X-axis represents the variable ranking threshold $N$ in a normalized form $\theta_N=\frac{N}{|\mathcal{K}_\ast|} \times 100\% $; for example, a data point at $x=31$ represents the result of a model that examines top-$31\%$ reconstructions or predictions, or equivalently top-9 out of $|\mathcal{K}_0|=31$ keys.  The Y-axis represents the $F_1$ score of the models.  In Figure~\ref{seqfig:logkeys31.png}, Baseline (5K) has its peak $F_1$ score $97.52\%$ at $\theta_N=31\%$, which is similar to the ones reported in the previous work~\cite{deeplog,nlsalog}; hence,  we believe Baseline (5K) and Baseline are representative models of predictor-based models. Besides DabLog and Baseline, we also include a trivial frequency model, which is mentioned in Section~\ref{seqsec:motivation}, for reference.  This trivial frequency model reports anomalous sequences by checking whether a sequence includes any event that is not an instance of top-$N$ most frequent keys.  It has its peak $F_1$ score $85.00\%$ at $\theta_N=29\%$.

%We are more interested in whether 
To assess the performance of these models at larger numbers of discrete event keys, we test them against $\mathcal{K}_1$ and $\mathcal{K}_2$.  Figure~\ref{seqfig:logkeys101.png} depicts the $F_1$ score trends of models upon $\mathcal{K}_1$.  DabLog (5K) has its peak $F_1$ score at $95.20\%$ at $\theta_N=27\%$, whereas Baseline (5K) has its peak $F_1$ Score $94.98\%$ at $\theta_N=31\%$; by comparing the trends, we can see that DabLog (5K) has a higher peak and a wider plateau, and hence DabLog (5K) is more advantageous for critics where coarse-grained reconstruction, say $N \geq 30\%$, is used.  Similarly, DabLog has its peak $F_1$ score $97.18\%$ at $\theta_N=9\%$, whereas Baseline has its peak $F_1$ Score $87.32\%$ at $\theta_N=10\%$; by comparing the trends again, we can see that DabLog is more advantageous for critics where fine-grained (i.e., more precise) reconstruction, say $N\leq10$, is used.  On the other hand, Figure~\ref{seqfig:logkeys304} shows a similar trends upon $\mathcal{K}_2$.  DabLog has its peak $F_1$ score $94.15\%$ at $\theta_N=6\%$, and Baseline also has its peak $F_1$ score at $\theta_N=6\%$ but its score is only $80.47\%$.  
The above comparisons clearly shows DabLog and DabLog (5K) are more advantageous upon all key sets $\mathcal{K}_0$, $\mathcal{K}_1$, and $\mathcal{K}_2$.

Unlike traditional deep neural networks, DabLog and Baseline does not follow the common belief of \textit{``having more training data leads to better performance''}, because their last components (i.e., anomaly critics) are not based on learning, but based on the ranks of keys regardless of training size.  This statement also applies to prior work~\cite{deeplog, nlsalog} that trained models with no more than $1\%$ of the dataset.  We can see in Figure~\ref{seqfig:logkeys31.png}, Figure~\ref{seqfig:logkeys101.png}, and Figure~\ref{seqfig:logkeys304} that the models with 5K training data have higher $F_1$ plateaus; however, it does not mean that the less training data, the better the models perform.  Rather, the less keys involved in training, the more keys have zero distribution (i.e., $p_i=0$) in reconstruction.  Since keys $k_i$ with $p_i=0$ have the same escalated rank, the anomaly critics tend to report them normal (i.e., hit) when a moderately high-threshold rank is applied.  Hence, the $F_1$ scores remain high until false-negative rises.  This is an unaddressed issue of the rank-based approach in prior work~\cite{deeplog, nlsalog}, and we believe that presenting results from DabLog (5K) and Baseline (5K) is misleading.  Models should always be trained by sufficient data, or otherwise the models learn very little about the keys.  Hence, we focus on results from DabLog and Baseline.

%Although DabLog and DabLog (5K) look good for both fine-grained reconstruction and coarse-grained reconstruction, we are more interested in critics where fine-grained reconstruction is required, because we want to narrow down \textit{``what are normal''} in hope of figuring out \textit{``why anomalies are abnormal''}, when we see anomalies.  
To understand the trend of $F_1$ score in DabLog, we depict the trends of \textit{different metrics} upon $\mathcal{K}_1$ in Figure~\ref{seqfig:dablog101.png}.  
%and DabLog upon $\mathcal{K}_2$, respectively; 
Again, X-axis represents the normalized variable ranking threshold $\theta_N$, and Y-axis represents the scores.  We can see that FP-rate are constantly low, and precision are constantly high while $F_1$ score decreases with $\theta_N$.  This is because when $\theta_N$ increases, more and more events (hence more sequences) will be considered as normal. In the extreme case $\theta_N=100\%$, everything will be considered normal. In other words, the number of FNs increases with $\theta_N$. This in turn causes recall to decrease, and accordingly $F_1$ score decreases with $\theta_N$.   
%recall decreases, and recall decreases because, as $\theta_N$ increases, the fixed rank $r_e>N$ of an abnormal event $e \in s_T$ can become $r_e \leq N'$, causing the models to wrongly consider $e$ normal.  This increasing number of FNs cause the decrease of recall. 

Figure~\ref{seqfig:precision.png} further shows that DabLog has much higher precision than Baseline in different parameter settings. This result is significant because a higher precision means fewer false positive cases for security administrators to (manually) analyze.    
Figure~\ref{seqfig:recall.png} shows that while the recall of DabLog and Baseline both decreases with $\theta_N$,
%depicts the trends of decreasing recall of different models upon $\mathcal{K}_1$ and $\mathcal{K}_2$, and we can see that 
DabLog is more advantageous as its trend decreases slower. This supports one of our insights mentioned in Section~\ref{seqsec:motivation}: since Baseline cannot identify structurally abnormal sequences, it would have more FNs (a FN case study is provided in the following subsection).

Previous work~\cite{deeplog, nlsalog} was also evaluated by the area under Receiver Operating Characteristic (ROC) curve in a 2-D space, where the X-axis represents the FP-rate, and the Y-axis represents the TP-rate.  Table~\ref{seqtab:auroc} lists the area under ROC of different models in our evaluation.  The area under ROC is useful when a model involves a variable threshold, as the curve depicts the expected trade-offs between TP-rate and FP-rate when tuning the variable threshold.  However, in our evaluation, the curves are almost right-angle lines, and the areas under ROC curves are very high.  That said, we can still see that DabLog and DabLog (5K) are more advantageous.

\begin{table}
\caption{Area Under ROC Curve}
\label{seqtab:auroc}
\centering
\begin{tabular}{rccc}
\toprule
& \multicolumn{3}{c}{Event Key Sets}\\
& $|\mathcal{K}_0|=31$ & $|\mathcal{K}_1|=101$ & $|\mathcal{K}_2|=304$ \\
\midrule
DabLog & 99.49\% & 99.44\% & 99.08\% \\
DabLog (5K) & 99.47\% & 99.34\% & 98.98\% \\
BaseLine & 96.02\% & 97.23\% & 97.41\% \\
BaseLine (5K) & 99.42\% & 99.29\% & 98.23\% \\
\bottomrule
\end{tabular}
\end{table}

\subsection{Case Study}

The above statements may seem too abstract and non-intuitive to understand why DabLog is more advantageous, so here we make more detailed comparisons under a fixed configuration.  
We motivate our case studies with this question: \textit{how does the fundamental difference make DabLog more advantageous?} 
Upon $\mathcal{K}_1$ and at $\theta_N=9\%$, DabLog and Baseline reported 14,768 common TPs, 627 common FPs, and 80 common FNs; however, DabLog also reported 1,986 exclusive TPs with trade-off 425 exclusive FPs, whereas Baseline reported 4 exclusive TPs but 2,215 exclusive FPs (recall that the testing dataset for $\mathcal{K}_1$ has 200,000 normal sessions and 16,868 abnormal sessions).  Upon $\mathcal{K}_2$ and at $\theta_N=6\%$, DabLog and Baseline reported 14,555 common TPs, 1,089 common FPs, and 90 common FNs; however, DabLog also reported 2,169 exclusive TPs with trade-off 932 exclusive FPs, whereas Baseline reported 24 exclusive TPs but 4,119 exclusive FPs.  In conclusion, DabLog provided more TPs and less FPs (or equivalently more TNs), and Baseline provided more FPs and less TPs (or equivalently more FNs).  We conducted case studies upon $\mathcal{K}_1$ and at $\theta_N=9\%$.  One was detailed in Section~\ref{seqsec:motivation}: a normal session that DabLog reported normal, while Baseline reported it abnormal.  

Here, we illustrate their difference through an opposite case.  DabLog reported an abnormal session (ID: -9134333392518302881) abnormal, while Baseline wrongly reported it normal.  It has in total 34 events, and DabLog reported the subsequences $s_{23}$, $s_{28}$, $s_{29}$, and $s_{30}$ abnormal, where $s_{23}=[e_{14}, \dots, e_{23}]$ and $s_{30}=[e_{21}, \dots, e_{30}]$.  The events are listed in Table~\ref{seqtab:tpfn}.  

These subsequences are considered abnormal because DabLog could not correctly reconstruct particularly the 21st event $e_{21}$.  That is, the key $k_3$ is not within the top-$9\%$ reconstructions for $e_{21}$.  Top-$9\%$ reconstructions include $k_4$, variants of $k_5$, variant of $k_6=$\textit{``addStoredBlock: blockMap updated ...''}, and $k_7=$ \textit{``EXCEPTION: block is not valid ...''}.  These event keys, except $k_7$, are frequent keys each dominates over $0.1\%$ of the dataset. Interestingly, here DabLog expects not only frequent keys, but also an extremely rare event key $k_7$ (which dominates $0.0017\%$) before $k_5$. Since these expected keys in top-$9\%$ reconstructions for $e_{21}$ are related to exception, verification, or blockMap updates, we believe that the reconstruction distribution is derived for causality relationship with $e_{23}$ (which is related block transmission) rather than for $e_{19}$ (which is related to block deletion), even though DabLog knows a deletion is asked at $e_{19}$ as it has correctly reconstructed $e_{19}$.  Our interpretation is that, DabLog expects a cause at $e_{21}$ that leads to the exception at $e_{23}$, and it is the absence of causality before $e_{23}$ making the sequence structurally abnormal.

In contrast, Baseline does not predict $e_{21}$ to be any of these keys after $s_{20}=[e_{11},\dots,e_{20}]$.  In other words, Baseline does not expect a cause at $e_{21}$, because it cannot foresee $e_{23}=k_5$.  With the fundamental limitation of unable to exploit bi-directional causality, Baseline is incapable of detecting such a structurally abnormal session.
%that is, even though Baseline can correctly predict $e_{23}=k_5$ for the corresponding subsequence $s_{22}$ ($k_5$ is a frequent key that dominates $1.81\%$), Baseline cannot foresee $k_5$ and expect particular causality before $k_5$.  
%In addition, Baseline thinks $e_{21}=k_3$ is normal, not only because $k_3$ (which dominates $0.53\%$) is not so rare, but also because the prior event $e_{19}=k_1$ escalates the probability of $k_3$.  
Therefore, we believe it is necessary for an anomaly detection methodology to see sequences as atomic instances and examine the bi-directional causality as well as the structure within a sequence.  Single-direction anomaly detection like Baseline cannot identify structurally abnormal sequences.  
\begin{table}
\caption{Example Sequential Discrete Events}
\label{seqtab:tpfn}
\centering
\begin{tabular}{ccl}
\midrule
$e_{14}$ & & Starting thread to transfer block\\
$e_{15}$ & & Receiving block within the localhost\\
$e_{16}$ & & blockMap updated: $10.251.\ast$ added of size 60-70 MB\\
$e_{17}$ & & Received block within the localhost \\
$e_{18}$ & & Transmitted block within the subnet\\
$e_{19}$ & $k_1$ & ask $10.251.\ast$ to delete block(s) \\
$e_{20}$ & $k_2$ & blockMap updated: $10.251.\ast$ added of size 60-70 MB\\
$e_{21}$ & $k_3$ & Deleting block /mnt/hadoop/dfs/data/current/...\\
$e_{22}$ & $k_4$ & Verification succeeded for block \\
$e_{23}$ & $k_5$ & Got exception while serving block within the subnet\\
$e_{24}$ & & Got exception while serving block within the subnet\\
$e_{25}$ & & Verification succeeded for block \\
$e_{26}$ & & delete block on $10.251.\ast$ \\
$e_{27}$ & & delete block on $10.251.\ast$ \\
$e_{28}$ & & delete block on $10.251.\ast$ \\
$e_{29}$ & & ask $10.251.\ast$ to delete block(s) \\
$e_{30}$ & & Deleting block /mnt/hadoop/dfs/data/current/... \\
\bottomrule
\end{tabular}
\end{table}

%The above demonstrates the reason why we prefer fine-grained reconstruction (e.g., $N \leq 10\%$) over coarse-grained reconstruction (e.g., $N \geq 30\%$): we want to narrow down \textit{``what are normal''} in order to figure out why \textit{``anomalies are abnormal''}.  In comparison, if coarse-grained reconstruction is applied, we will suffer from simultaneously analyzing more than 30 normal event keys.

}

    {\section{Discussion and Future Work} \label{seqsec:discussion}

\subsection{A More Comprehensive Embedding} \label{seq:discussion:embedding}

Our DabLog approach learns the embedding function $\mathcal{E}$ by training an additional embedding layer along with the other layers.  This approach has the drawback of handling \textit{unknown event keys} that are not in the training data but in the testing data. If $e_\ast=k_\ast$ is not in the training data, then $x_\ast=\mathcal{E}(k_\ast)$ is an undefined vector; hence the model may wrongly judge sequences $S_\ast$ that includes $k_\ast$.  In the literature of natural language processing, this problem is also known as the \textit{out-of-vocabulary} problem, and there are two common workaround options for it.  One option is to substitute the designated \textit{``unknown''} word for not only unknown words but also rare words, so that \textit{``unknown''} is trained as if it is some known rare events.  This option, however, is not applicable to security audit logs, because unknown events can be frequent and similar to known events (e.g., unknown event key \textit{``receiving 70-80 MB''} is similar to known event key \textit{``receiving 60-70 MB''}), and wrongly treating unknown events as rare events may cause FPs.  The other option is to leverage a pre-trained Word2Vec embedding in building a new embedding model (e.g. Mimick embeding~\cite{mimick}).  Inspired by the latter option, one of our future work is to build an Event2Vec embedding model $\mathcal{E}'$ that can derive the embedding for $k_\ast$ by examining the words in $k_\ast$.  We will investigate how this new embedding can improve the results in our future work.

%To tackle unknown events, DabLog has a more comprehensive Event2Vec option $\mathcal{E}'$ that combines a pre-trained Word2Vec embedding and the training of a customized Mimick embedding (Figure~\ref{seqfig:mimick.pdf}).  Our $\mathcal{E'}$ takes a discrete event $e_t$ as input, splits $e_t$ into words $w_1$ to $w_n$, and then feeds these words into a customized Mimick network (i.e., bi-directional LSTM and deep perceptron).  To train $\mathcal{E'}$, a pre-train phase for $\mathcal{E}$ is required. In this paper, we did not detail $\mathcal{E}'$ in the main body, because the dataset we use (in which there are only have 32 unique events, and unknown events are certainly abnormal) does not need Mimick, and hence currently we cannot conduct a comprehensive evaluation about how Mimick can improve the models.  We leave it as a future work.

%\seqfig{mimick.pdf}{0.8}{DabLog with Word2Vec and Mimick Embedding}

\subsection{A Rank-Distance Double-Threshold Criterion} \label{seq:discussion:criterion}

We presented DabLog's critic as a rank-based criterion in order to compare DabLog with existing work~\cite{deeplog, nlsalog}.  However, both rank-based and threshold-based criteria have the same major drawback: they brutally divide the probabilistic distribution $P_\tau$ into two parts (normal and abnormal), while omitting the correlation between the true $\hat{k}_i$ and the other keys $\{k_j\}$.  The problem is that the critic may wrongly see a normal event key as abnormal, and vice versa.  For example, let us consider a case in which $| p_i - p_j |$ is very small, meaning that the corresponding keys $\hat{k}_i$ and $k_j$ are almost equivalently anomalous to the model, still $\hat{k}_i$ could be abnormal and $k_j$ could be normal when the pivotal condition sits in between.  If $\hat{k}_i$ has certain strong correlation with another $k_J$ (e.g., neighbors in embedding universe $\mathcal{U}$), then very likely $\hat{k}_i$ has the same anomaly label as $k_J$.  We believe both rank-based criterion and threshold-based criterion have limitation that cause FPs and FNs.  Therefore, we are curious  whether we can incorporate the embedding distance in a rank-based critic (we name the resulting criterion \textit{rank-distance double-threshold criterion}).  Basically, in addition to checking top-$N$ reconstructions, it also checks the labels of neighbors (within threshold radius) in $\mathcal{U}$.  We leave it as a future work.

\begin{table}
\caption{Experiments upon $\mathcal{K}_1$ and with $\theta_N=7.5\%$}
\label{seqtab:seqlen}
\centering
\begin{tabular}{rrrrr}
\toprule
& seqlen $=10$ & seqlen $=30$ & Intersection & Union \\
\midrule
TP & 16,367 & 14,910 & 14,630 & 16,647 \\ 
FP & 2,424 & 1,224 & 1,059 & 2,589 \\
TN & 197,576 & 198,776 & 198,941 & 197,411 \\
FN & 471 & 1,928 & 2,208 & 191 \\
FP Rate & 1.21\% & 0.61\% & 0.52\% & 1.29\% \\
Recall & 97.20\% & 88.54\% & 86.88\% & 98.86\% \\
Precision & 87.10\% & 92.41\% & 93.25\% & 86.54\% \\
$F_1$ Score & 91.87\% & 90.44\% & 89.95\% & 92.29\% \\
\bottomrule
\end{tabular}
\end{table}

\subsection{Merging Models of Different Ideas}
There are many different ideas on how to build an anomaly detection model, and each has its advantages.  In hope of being more advantageous, one may want to merge different ideas by merging their anomaly results according to certain rules (e.g., intersection or union), and we refer to the resulting method as a \textit{hybrid model}.  From Table~\ref{seqtab:seqlen}, we can see that taking intersection benefits us by less FPs, and taking union benefits us by less FNs.  Besides hybrid models, we can also consider building a \textit{composite model}, which is an interconnected neural network resulted from merging the neural networks of different ideas (different sub-networks are trained simultaneously).  We have many options for merging models, yet \textit{``which advantage} (e.g., having less FPs or less FNs) \textit{is more preferable''} is debatable. 
In practice, having less FPs is more preferable for offline learning models, as manual inspections are very expensive.  In contrast, having less FNs is more preferable for online learning models that are capable of unlearning FPs~\cite{lifelong}.  Among these different options, although we believe that there is little space for improvement, their benefits are worth researching into in the future.

\subsection{DabLog is beyond a Standard Autoencoder}

We would like emphasize again that, although DabLog is based on the autoencoder methodology, DabLog is more than a standard autoencoder.  
Compared to the reconstruction problem for standard autoencoders, our problem---\textit{identifying time-sensitive anomaly by examining discrete events}---is more like a language processing problem.
We can view sequences $S$ as sentences and events $e_t$ as words, and we care more about wording (e.g., \textit{``which words $e_t$ better fit in the current sentence $S$''}) rather than embedding (e.g., \textit{``which vector $y_\tau$ better vectorize $e_t$ in the current sentence $S$''}).  
As such, DabLog is designed as an \textit{embed-encode-decode-classify-critic} model, so that it can help us with finding \textit{``fitting words''} and \textit{``unfitting words''}, or equivalently, normal events and abnormal events in the sequential context of $S$.
In contrast, typical time-insensitive autoencoder-based anomaly detection methods directly use scalar reconstruction errors (e.g., root-mean-square error) as anomaly scores.  

\subsection{More Reconstruction Methods and More Datasets}

Although we show that reconstructing sequences is also an attractive methodology (besides predictor-based methods), autoencoders are not the only sequence-reconstruction solution.  Combining predictor results from predictors of different directions can also achieve sequence reconstruction (i.e., with one for successor events and the other for predecessor events), though this approach may not be as accurate or efficient as the autoencoder-based approaches. Nevertheless, the performance difference of  different sequence-reconstruction methods is worth researching into. We will also put more efforts in discovering and experimenting more datasets. In this paper we use the same HDFS dataset as used in the prior work for  a convenient comparison; however, this dataset has little to do with cyberattacks or cyber threats. Although we are aware of some other threat-related datasets~\cite{cert, lanl}, they are too abstract (either because low-level events are not included or because too much details are discarded for anonymity) to learn sequential relationships between events. As a consequence, their sequences are not reconstructable, unless domain knowledge is available for numeric feature extraction, as shown in Liu et.al~\cite{liuliu1,liuliu2}'s work (but then the problem becomes numeric-feature reconstruction and is no longer discrete-key sequence reconstruction).  DabLog requires datasets that include relationships among low-level events (e.g., system-call events or Windows audit events with details).  To be more attractive to the security community, our top priority in our future work is to find more (or engineer) labeled datasets for experiments.}

    {\section{Conclusion} \label{seqsec:conclusion}

With regard to anomaly detection approaches for discrete events, we address a fundamental limitation of the widely adopted predictor-based methodology through our in-depth case studies.
We argue that recomposing sequences is also an attractive methodology, especially in real-world contexts where the need of detection with more keys cannot be well satisfied by predctor-based methodology.
We propose DabLog and evaluate DabLog with the HDFS console-log dataset.
Our results show that DabLog outperforms our predictor-based baseline model in terms of $F_1$ score.
With reconstruction of sequential events, not only does DabLog have much fewer FPs, but also does DabLog improve awareness regarding \textit{what is normal} in contexts that involves more keys.
}

%\section*{Acknowledgment}

\bibliographystyle{IEEEtranS}
\bibliography{reference}

\end{document}